\documentclass[runningheads]{llncs}

\usepackage{eccv}

\usepackage{eccvabbrv}

\usepackage{graphicx}
\usepackage{booktabs}
\usepackage{arydshln}
\usepackage{booktabs}
\usepackage{enumitem}
\usepackage{balance}
\usepackage{combelow}
\usepackage{tabularx}
\usepackage{cite}
\usepackage{comment}

\usepackage{wrapfig}
\usepackage[stable]{footmisc}

\usepackage{multirow}
\usepackage{makecell}
\usepackage{mathtools}
\usepackage{algorithm}
\usepackage{graphicx}
\usepackage{algorithmicx}
\usepackage{eqparbox,array}

\usepackage{xcolor,colortbl}

\usepackage{subcaption}
\usepackage{caption}

\usepackage{setspace}
\usepackage{url}

\usepackage{mwe} 
\usepackage{calc} 
\usepackage{pifont} 
\newcommand{\cmark}{\color{blue}{\ding{51}}}
\newcommand{\xmark}{\color{red}{\ding{55}}}

\newcommand{\redbox}{\raisebox{-1pt}{\includegraphics[height=9px]{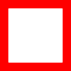}}}
\newcommand{\orangebox}{\raisebox{-1pt}{\includegraphics[height=9px]{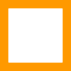}}}
\newcommand{\violetbox}{\raisebox{-1pt}{\includegraphics[height=9px]{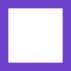}}}

\usepackage[accsupp]{axessibility}  

\usepackage{hyperref}

\usepackage{orcidlink}

\begin{document}

\newcommand{\method}{{\textsc{ReALFRED}}\xspace}
\newcommand{\methodfull}{{\textsc{Real-World ALFRED}}\xspace}
\title{\method: An Embodied Instruction Following Benchmark in Photo-Realistic Environments}

\titlerunning{\method}

\author{Taewoong Kim\inst{1, *}\orcidlink{0000-0001-9999-6474} \and
Cheolhong Min\inst{1, *}\orcidlink{0009-0007-1495-2976} \and
Byeonghwi Kim\inst{1}\orcidlink{0000-0003-3775-2778}  \and \\
Jinyeon Kim\inst{1, 2}\orcidlink{0000-0001-9122-3194}  \and
Wonje Jeung\inst{1}\orcidlink{0009-0008-8975-5720}  \and
Jonghyun Choi\inst{1, \dagger}\orcidlink{0000-0002-7934-8434}}

\authorrunning{T.~Kim and C.~Min et al.}


\institute{$^1$ Seoul National University $^2$ Yonsei University\\
\email{\{twoongg.kim, cheolhong.min, byeonghwikim\}@snu.ac.kr, jinyeonkim@yonsei.ac.kr, \{specific0924, jonghyunchoi\}@snu.ac.kr}}

\maketitle

\begin{abstract}
    Simulated virtual environments have been widely used to learn robotic agents that perform daily household tasks.
    These environments encourage research progress by far, but often provide limited object interactability, visual appearance different from real-world environments, or relatively smaller environment sizes. 
    This prevents the learned models in the virtual scenes from being readily deployable.
    To bridge the gap between these learning environments and deploying (\ie, real) environments, we propose the \method benchmark that employs real-world scenes, objects, and room layouts to learn agents to complete household tasks by understanding free-form language instructions and interacting with objects in large, multi-room and 3D-captured scenes.
    Specifically, we extend the ALFRED benchmark with updates for larger environmental spaces with smaller visual domain gaps.
    With \method, we analyze previously crafted methods for the ALFRED benchmark and observe that they consistently yield lower performance in all metrics, encouraging the community to develop methods in more realistic environments.
    Our code and data are publicly available\footnote[3]{Homepage: \url{https://github.com/snumprlab/realfred}}.
\end{abstract}

\def\thefootnote{*}\footnotetext{Equal contribution. $^\dagger$Corresponding author.}

\section{Introduction}
Building autonomous robotic assistants that can perform everyday household tasks has been an elusive aspiration within the research community for decades.
To let them learn these intricate tasks, we may provide them with interactive environments where agents can learn task completion skills with numerous interactions with environments. 
A straightforward approach to train such agents that can carry out real-world activities is to directly deploy robots in real-world environments and let them learn to complete desired tasks.
However, this often faces several practical challenges, including cost, time, or safety concerns~\cite{Jie2018sim2real,kumar2021rma,2018-ICLR-distill,science2023quadrupedal}.

\begin{figure}[t!]
    \centering
    \includegraphics[width=.9\linewidth]{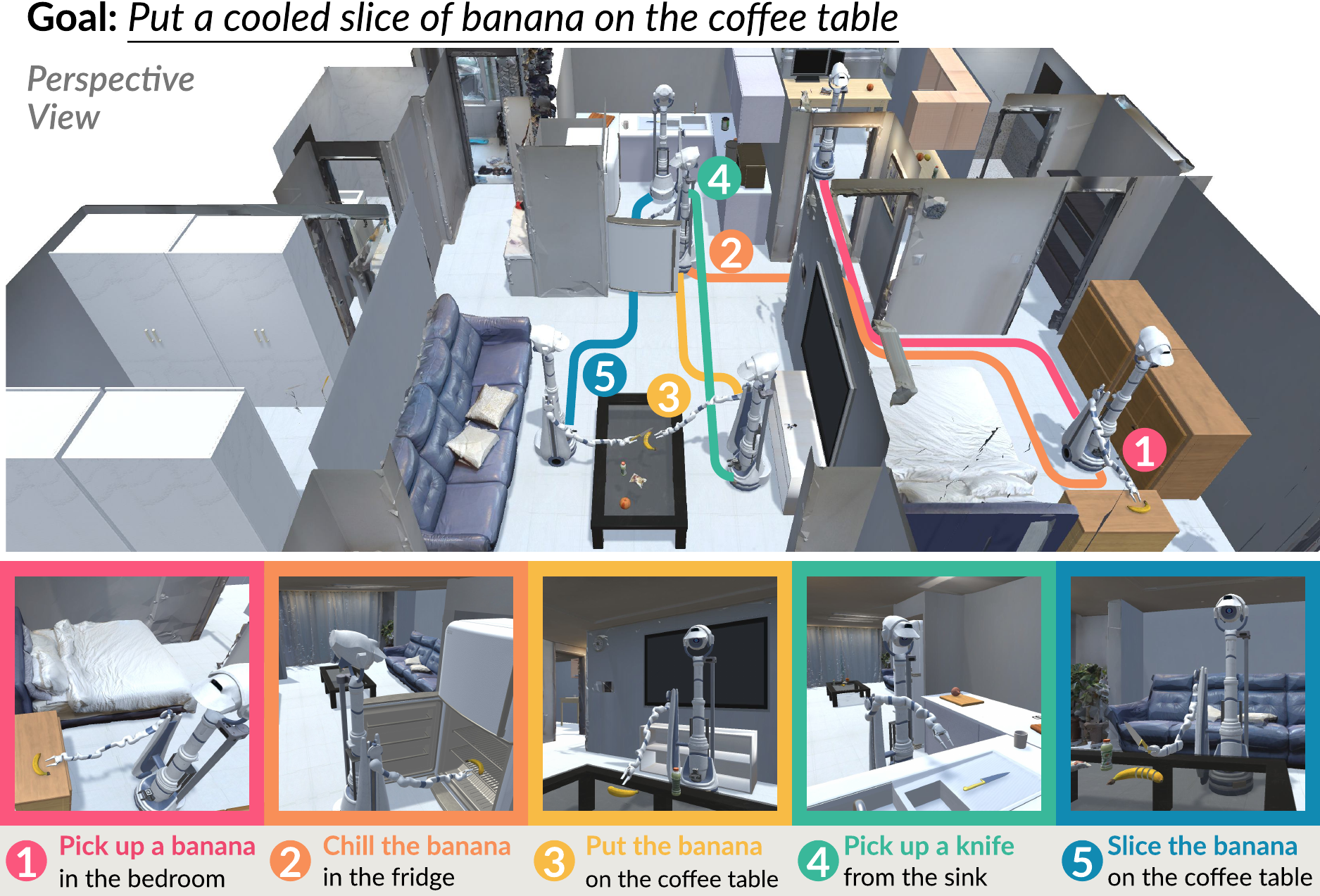}
    \caption{
        \textbf{Proposed \method benchmark.}
        The top image provides a perspective view of one of our scenes.
        The images below represent third-person views at each time step, along with their corresponding descriptions, for better understanding.
        The agent is required to understand instructions in natural language and then complete the desired tasks by navigating large 3D-captured environments and interacting with objects.
    }
    \label{fig:main_fig}
\end{figure}

As an alternative, several simulated environments have been introduced~\cite{habitat19iccv, xia2018gibson} which leverage extensive 3D-captured environments obtained from real-world scenes~\cite{chang2017matterport3d, ramakrishnan2021hm3d}.
Compared to real-world deployment, these environments offer agents a faster process of taking actions and observing consequences, and the convenience of resetting the environment and trying again in case of failure, which enables agents to learn the skills to complete desired tasks.
Adopting such simulated environments has produced remarkable advancements in various subtasks for embodied AI agents, including visual navigation~\cite{chaplot2020object, min2023object, ramakrishnan2022poni, wijmans2019dd, Partsey_2022_CVPR}, vision and language navigation~\cite{anderson2018vision,krantz2020navgraph}, and remote object grounding~\cite{qi2020reverie}.
Due to the inherently static nature of these 3D-captured environments where objects (\eg, books, chairs, \etc) remain non-interactive, however, current benchmarks~\cite{chaplot2020object, anderson2018vision} for these tasks have less focused on object interaction, which might hinder deployability for more complex tasks that require object interaction.
Recent studies~\cite{ramrakhya2022habitat, majumdar23findthis_short} insert liftable objects in scanned environments for object interaction, but they support limited object interaction such as picking up objects, which might not provide enough deployability for more challenging real-world scenarios such as heating objects using a microwave or cooling objects using a refrigerator.

Meanwhile, virtual game engines such as Unity have been exploited to build object-interactable environments with graphically crafted assets, including walls, floors, ceilings, and objects.
These object-interactable environments have led to notable progress in the handling of more intricate object-centric tasks, including rearrangement~\cite{szot2021habitat,weihs2021visual} and manipulation~\cite{gu2023maniskill2, heo2023furniturebench}, beyond navigation-centric tasks.
In particular, we have observed significant progress in the execution of more complex tasks by natural language instructions~\cite{shridhar2020alfred, singh2021factorizing, pashevich2021episodic, kim2021agent, blukis2021persistent, min2021film}.
However, object-interactable environments for training and evaluation of such agents often pose several issues, such as visual domain gaps~\cite{tobin2017domain} and relatively smaller room sizes compared to their counterparts in 3D-captured environments~\cite{ramakrishnan2021hm3d, chang2017matterport3d, straub2019replica}.

\begin{wrapfigure}{r}{.45\textwidth}
    \centering
    \includegraphics[width=0.9\linewidth]{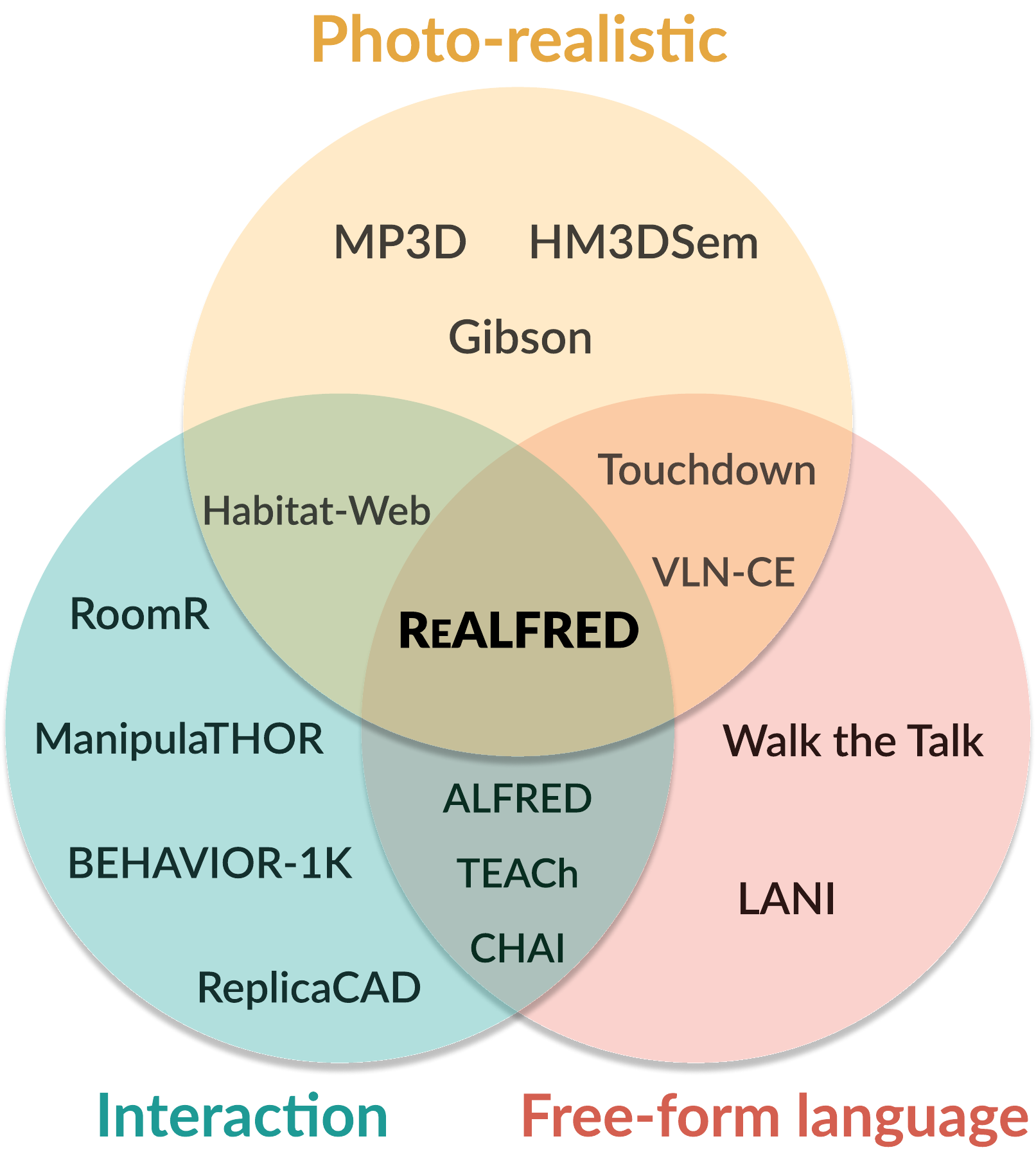}
    \caption{
         While other benchmarks~\cite{chang2017matterport3d, ramakrishnan2021hm3d, xia2018gibson, ramrakhya2022habitat, chen2019touchdown, krantz2020navgraph, weihs2021visual, ehsani2021manipulathor, szot2021habitat, shridhar2020alfred, padmakumar2022teach, macmahon2006walk, misra2018mapping, li2022behavior} provide one or two aspects, our proposed \method benchmark addresses all of these aspects.
    }
     \label{fig:Diagram}
\end{wrapfigure}

To bridge the gap between limited object interactability and environmental sizes with visual domain discrepancy, we propose the \methodfull (\method) benchmark that requires agents to complete long-horizon tasks by understanding free-form language instructions and interacting with objects in large 3D-captured environments, following a similar task setup to the widely used embodied instruction following benchmark, ALFRED~\cite{shridhar2020alfred}.
The 3D-captured environments used in the \method benchmark encompass multiple rooms, providing ample space for agents to engage in multiple rooms in a single episode, which adds a sense of realism to the tasks.
This resembles real-world scenarios in which agents navigate seamlessly between multiple rooms.
Unlike prior benchmarks focusing primarily on single-room activities~\cite{shridhar2020alfred} or 3D-captured environments with limited object interaction~\cite{habitat19iccv}, \method provides task evaluation in wide, realistic, and interactive environments to mirror the natural expectations of human-robot interactions.
Fig.~\ref{fig:main_fig}. illustrates the household task in one of the scenes in ours.

In our experiments, we observe that the models~\cite{shridhar2020alfred, singh2021factorizing, kim2021agent, min2021film, blukis2021persistent, song2023llmplanner} proposed in synthetic environments~\cite{shridhar2020alfred} do not perform well in our \method benchmark, implying that models developed for synthetic environments may not easily adapt to realistic environments.

\noindent We summarize our contributions as follows:
\begin{itemize}[leftmargin=10pt]
    \item We propose \method, a benchmark for embodied instruction following with 3D-captured multi-room environments and objects.
    \item We collect 3D-captured scenes and objects to reduce the simulation-reality gap and free-form language instructions to support task completion based on agents' language understanding.
    \item We provide analyses on the recent state-of-the-art models in the literature and the relevant Sim2Real transfer to empirically validate the necessity of our \method benchmark.
\end{itemize}


\section{Related Work}
Fig.~\ref{fig:Diagram} illustrates the comparison of our \method with other benchmarks in three selected aspects.
We first review 3D-captured environments and benchmarks that provide visual aesthetics similar to real-world environments.
Then, we review simulation environments and benchmarks that support object interaction.
Table~\ref{tab:Comparison} compares our \method with other indoor datasets.

\begin{table*}[t!]
    \caption{
        \textbf{Comparison of \method and other Embodied AI benchmarks.}
        `Language' column denotes the number of human annotated language directives.
        `Environment' column compares spatial characteristics and whether it supports interactivity. 
        `Inference' column denotes whether their action space includes interactive capability with objects.
        The \method benchmark is the first benchmark that provides a 3D-captured and interactable environment to solve household tasks that require navigation and interaction at the same time directed by natural language commands.
        $^\dagger$We count the number of dialogue sessions.
        $^\ddagger$We count the number of annotations in English.
        $^\ast$Though~\cite{deitke2020robothor} environment supports interaction, interaction is not required.
    }
    \centering
        \resizebox{\textwidth}{!}{    
        \begin{tabular}{lccccccc}
        \toprule
        \multirow{3}{*}{}                           & {Language}     & \multicolumn{4}{c}{{Environment}}      & {Inference} \\
        \cmidrule(lr){2-2} \cmidrule(lr){3-6} \cmidrule(lr){7-7}
                                                    & Human                 & Visual                & Multi-Room               & Movable                 & State                    & Object \\
                                                    & Annotations           & Quality               & Navigation               & Objects                 & Changes                  & Interactability \\
        \midrule
        IQA~\cite{gordon2018iqa}                    & -                     & \color{red}Synthetic & \xmark                             & \xmark            & \cmark                    & \cmark                      \\
        ManipulaTHOR~\cite{ehsani2021manipulathor};
        RoomR~\cite{weihs2021visual}                 & -                    &\color{red}Synthetic  &   \xmark                        & \cmark             & \cmark        & \cmark    \\  
        RoboTHOR~\cite{deitke2020robothor}           & -                     & \color{red}Synthetic & \cmark                             & \cmark            & \cmark                    & \xmark$^\ast$  \\
        ProcTHOR~\cite{deitke2022️}                   & -                    & \color{red}Synthetic & \cmark                             & \cmark            & \cmark                     & \cmark  \\
        ReplicaCAD~\cite{szot2021habitat}
            & -                    &\color{red}Synthetic  &  \cmark                          & \cmark           & \cmark       & \cmark    \\
        BEHAVIOR-1K~\cite{li2022behavior}; HSSD-200~\cite{khanna2024habitat}
            & -                    &\color{red}Synthetic  &  \cmark                          & \cmark           & \cmark       & \cmark    \\
        OpenRooms~\cite{Li_2021_CVPR}                & -                     & \color{blue}Photo     & \xmark                                 & \xmark            & \xmark                    & \xmark  \\
        MP3D~\cite{chang2017matterport3d};
        HM3D~\cite{ramakrishnan2021hm3d};
        Gibson~\cite{xia2018gibson}                  & -                     & \color{blue}Photo     & \cmark                         & \xmark              & \xmark                    & \xmark  \\
        Habitat-Web~\cite{ramrakhya2022habitat}                  & -     & \color{blue}Photo  & \cmark                               & \cmark              & \cmark                     & \cmark   \\
        ALFRED\cite{shridhar2020alfred}; TEACh~\cite{padmakumar2022teach};
        CHAI~\cite{misra2018mapping}             & 25k+; 2.0k+$^\dagger$;12k+           & \color{red}Synthetic & \xmark   & \cmark    & \cmark       & \cmark    \\
        LANI~\cite{misra2018mapping} ; Walk the Talk~\cite{macmahon2006walk}           & 28k+; 0.7k+                    & \color{red}Synthetic      & \cmark              & \xmark             & \xmark             & \xmark    \\
        Virtualhome~\cite{puig2018virtualhome}                                         & 2.7k+                          & \color{red}Synthetic      & \cmark              & \xmark             & \xmark             & \cmark    \\
        R2R~\cite{anderson2018vision}; RxR~\cite{rxr}; VLN-CE~\cite{krantz2020navgraph}           & 21k+; 42k+$^\ddagger$; 21k+              & \color{blue}Photo         & \cmark              & \xmark             & \xmark             & \xmark    \\
                      
        \midrule                             
        \textbf{\method}                            & \textbf{30k+}         & \textbf{\color{blue}Photo} & \cmark             & \textbf{\cmark}     & \textbf{\cmark}           & \textbf{\cmark}    \\
        \bottomrule
        \end{tabular}
    }    
    \label{tab:Comparison}
\end{table*}

\noindent \textbf{Datasets and benchmarks with 3D-captured environments.}
The collection of advanced 3D scans has played a key role in enhancing our understanding of 3D objects~\cite{song2015sun, dai2017scannet, yeshwanth2023scannet++} and their perception~\cite{hua2016scenenn, mao2022multiscan, dehghan2021arkitscenes}.
While these datasets offer valuable insights for a deeper understanding of 3D environments, they lack object interaction for learning interactive embodied agents.
To further promote research on embodied agents for real-world applications, training and evaluation of such agents with physical spaces from scanned data~\cite{chang2017matterport3d,xia2018gibson,straub2019replica} have been proposed.
They provide a rich source of data for researchers to explore the capabilities of agents operating within real-world-inspired scenarios.

In these environments, notable progress has been achieved, primarily in the realms of navigation and exploration.
Extensive research has been conducted on agents capable of navigating complex 3D environments, as evidenced by work such as navigating to a specified object~\cite{chaplot2020object, min2023object, ramakrishnan2022poni}, navigating to a certain point~\cite{wijmans2019dd, Partsey_2022_CVPR}, and navigating to the shown image~\cite{krantz2023navigating}.
Similarly, exploring novel environments has also yielded valuable insights~\cite{chaplot2020learning,chen2018learning}.

Additionally, there is a work to integrate multi-modal sensory information for developing an agent that can deeply understand environments using inputs such as vision and language~\cite{anderson2018vision, rxr} and audio and vision~\cite{chen2020soundspaces, chen22soundspaces2}.
These advances enable agents to utilize a broader range of sensory data, enhancing their perception and interaction capabilities. 
Further research explores additional capabilities of embodied agents, such as object localization alongside navigation~\cite{qi2020reverie}.

However, a fundamental limitation of these scan-based environments remains their static nature, which restricts the potential for object interaction.
This limitation has resulted in the majority of prior research primarily focusing on navigation tasks, leaving the broader scope of agent adaptability to more complex, interactive challenges largely unaddressed.
Consequently, this limitation hampers the adaptability of agents to more complex tasks.

\noindent \textbf{Object-interactable simulated environments.}
Scan-based environments~\cite{yeshwanth2023scannet++, ramakrishnan2021hm3d} are typically composed of static scene representations, focusing on the visual aspects of real-world spaces.
However, they often fail to adequately support interactivity with objects, representing a discrepancy between the dynamic and interactive characteristics of the real world, necessitating the development of environments better aligned with the demands of interactive agent research.

To address this gap, various simulators, such as AI2-THOR~\cite{ai2thor}, ManipulaTHOR~\cite{ehsani2021manipulathor}, RoboTHOR~\cite{deitke2020robothor}, VirtualHome~\cite{puig2018virtualhome}, TDW~\cite{gan1threedworld}, iGibson~\cite{li2021igibson}, and OmniGibson~\cite{li2022behavior}, have emerged as promising solutions.
These environments are engineered with a primary focus on enabling interaction between agents and objects.
They are built upon game engines, which provide a solid foundation for ensuring realistic interactivity within the virtual environment.
On these simulators, researchers publish benchmarks~\cite{shridhar2020alfred, padmakumar2022teach, deitke2020robothor, gao2022dialfred, srivastava2021behavior, li2022behavior, yenamandra2023homerobot, puig2023habitat} that support interactions.
These benchmarks have been promoting research in the field, leading to the development of robotic assistants capable of handling complex tasks.

However, the dataset that provide interactive environments are limited in size of a room because creating a large high-fidelity space is challenging~\cite{ramakrishnan2021hm3d}.
Furthermore, despite their near-photo realism, agents would face a visual domain gap when deployed in real world environments~\cite{zhang2019vr, truong2021bi, tobin2017domain}.
To this end, Habitat-Web~\cite{ramrakhya2022habitat} proposes an interactive pick-and-place task based on the Habitat simulator~\cite{habitat19iccv, szot2021habitat}.
Their template-based language instructions, however, might not be sufficient to express the complex nuances of human expression.
Innovative methods for acquiring 3D scans from phone-captured layouts have been proposed for learning enviroment-specific policies~\cite{deitke2023phone2proc}.
Nevertheless, its synthetic assets may lead to visual domain gaps when deployed in the real world.
On the contrary, our \method supports photorealism, high interactivity with objects, and free-form language annotations.
These features can offer a framework for developing language-driven agents with visual perception for complex household chores.

\section{The \method Benchmark}
To develop agents capable of performing household tasks, substantial progress has been achieved in various domains, including navigation~\cite{rxr, anderson2018vision}, rearrangement~\cite{weihs2021visual}, and manipulation tasks~\cite{gu2023maniskill2, heo2023furniturebench}.
In particular, \cite{shridhar2020alfred} recently introduced the ALFRED benchmark that requires agents to complete long-horizon household tasks by jointly understanding egocentric visual observations and natural language instructions in household environments.

\begin{figure}[t!]
    \centering
    \includegraphics[width=0.9\linewidth]{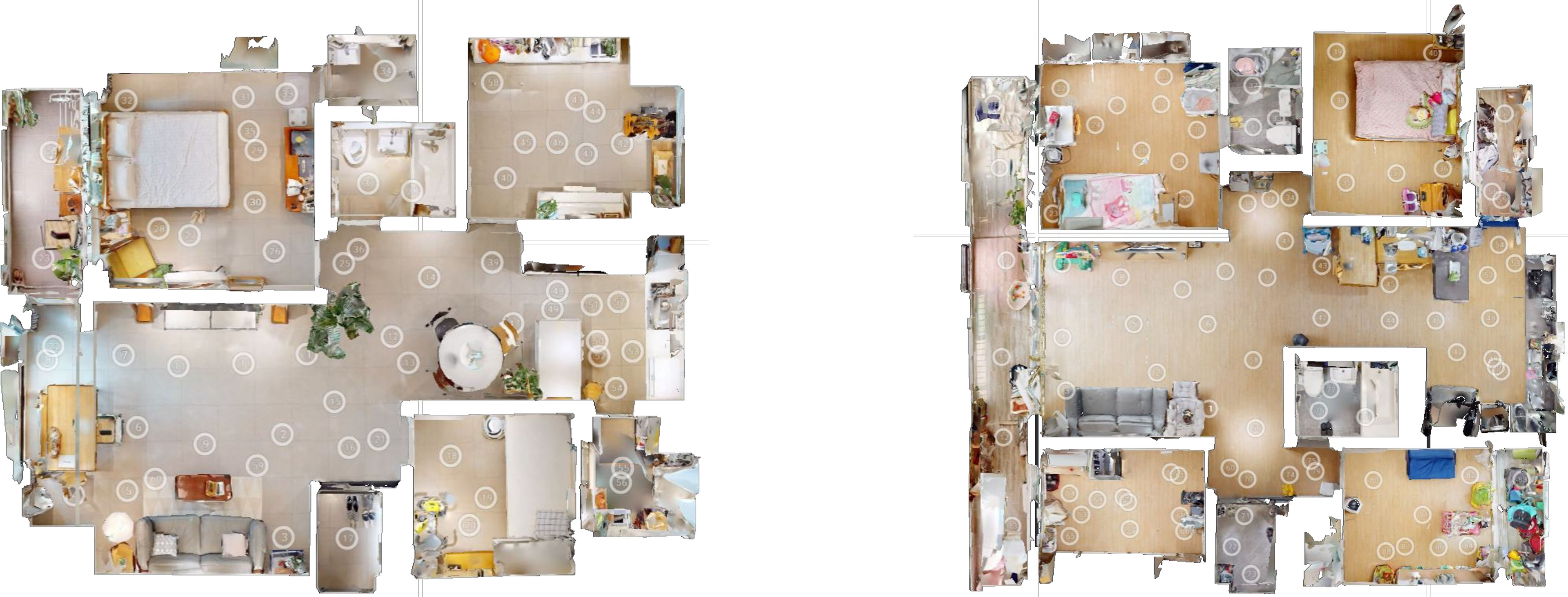}
    \caption{
        \textbf{Top-down view of 3D-captured environments.}
        We provide two examples from our scanned indoor environments.
        White circles denote where scanners are deployed.
        By scanning scenes at diverse points, we can prevent including blind spots.
    }
    \label{fig:scan_point}
\end{figure}

However, these environments are restricted to a single room size compared to previously proposed 3D-captured environments~\cite{ramakrishnan2021hm3d, chang2017matterport3d} consisting of multiple rooms, which could potentially restrict the deployability of agents to larger environments.
Furthermore, the environments used in the ALFRED benchmark~\cite{shridhar2020alfred} are built with synthetic CAD assets and therefore could potentially yield visual aesthetics different from those obtained from real-world environments~\cite{tobin2017domain}, which could eventually cause performance degradation due to visual domain gaps.

To address these issues, we extend the ALFRED benchmark~\cite{shridhar2020alfred} and propose a challenging benchmark, named the \method benchmark, which requires agents to perform household tasks in large indoor environments captured in 3D with object interaction.
For training and evaluation, we follow the same protocol as \cite{shridhar2020alfred} to collect expert demonstrations in the captured large environments.

\subsection{Object-Interactable 3D-Captured Scenes}
To reduce the visual domain gap, a straightforward approach is to use 3D scans of real world environments.
However, the captured 3D scans (\ie, meshes) remain \emph{static} and thus, agents cannot interact with objects in the captured scenes.
For object interaction, we manually replace object parts with 3D object assets to support object interaction.
For photorealism comparison with previous environments using FID~\cite{heusel2017gans} and KID~\cite{binkowski2018demystifying} metrics, please refer to Sec A.3.

We detail the process of collecting object-interactable 3D-captured environments and highlight key differences from previously proposed benchmarks below.

\noindent \textbf{Data acquisition process.}
To collect 3D scans of real-world environments, we visit residential properties and employ scanners.
Inspired by recent work~\cite{dehghan2021arkitscenes}, we collect scans outside of the US to add the diversity with public scanned environments.
We utilize the same 3D scanner as \cite{chang2017matterport3d}, equipped with three RGB cameras along with a depth sensor, and capture images from three distinct perspectives: front-facing and slightly vertical above and below.
Panoramic images are acquired through six consecutive captures with horizontal rotation from a fixed viewpoint.
We scan each house with $2.5$-meter intervals and address blind spots due to furniture with additional scans from different viewpoints.
Fig.~\ref{fig:scan_point} shows houses where scanners were deployed while capturing scans.
It can be discerned that captures were densely concentrated in areas with a high presence of objects.

\noindent \textbf{Object-interactable environments.}
To construct an environment with numerous interactable objects, separating each objects should be preceded.
In other words, objects in the scans are initially merged into the background and therefore they remain as the background. 
Constructing an environment where objects can be interacted with requires the separation of objects from the background and from each other.
Therefore, we manually separate the 3D scans into background elements and interactive objects.
Furthermore, each object can exist in various states.
To visualize changes in an object's state, we add state-relevant textures on objects.
For example, we add a stain texture to a clean object when it becomes dirty.
Finally, we reconstruct these individual object meshes within the Unity editor, making them compatible with the AI2-THOR simulator~\cite{ai2thor}.

\noindent \textbf{Comparison with previous benchmarks.}
\label{sec:comparision_with_prior_benchmark}
To investigate the spatial characteristics of scenes in the \method benchmark, we compare ours with other benchmarks that support interaction with objects~\cite{deitke2020robothor, shridhar2020alfred, szot2021habitat}.
We observe enhancements in our dataset in both: 1) spatial sizes and 2) spatial complexity.

\begin{figure}[t!]
    \begin{subfigure}{0.48\linewidth}
        \centering
        \includegraphics[width=\linewidth]{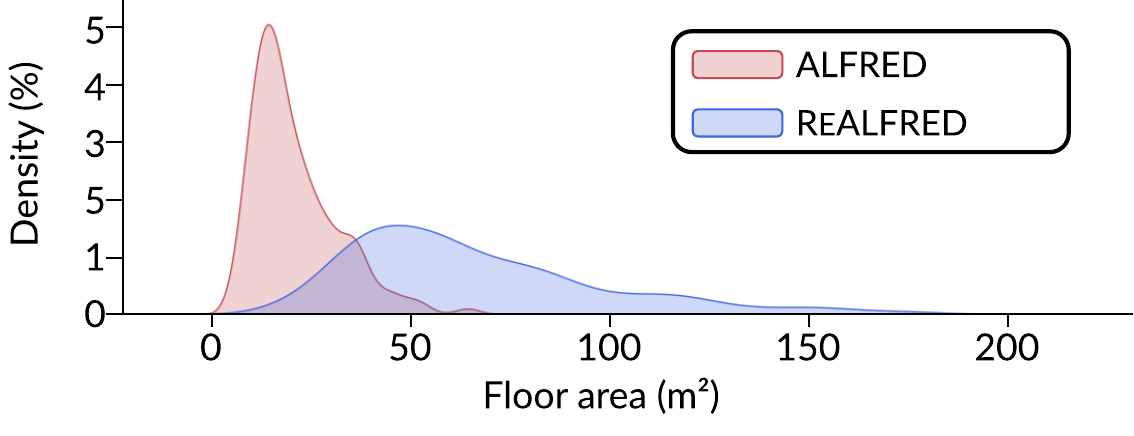}
        \caption{Distribution of floor areas by scene.}
        \label{fig:floor_area}
    \end{subfigure}
    \begin{subfigure}{0.48\linewidth}
        \centering
        \includegraphics[width=\linewidth]{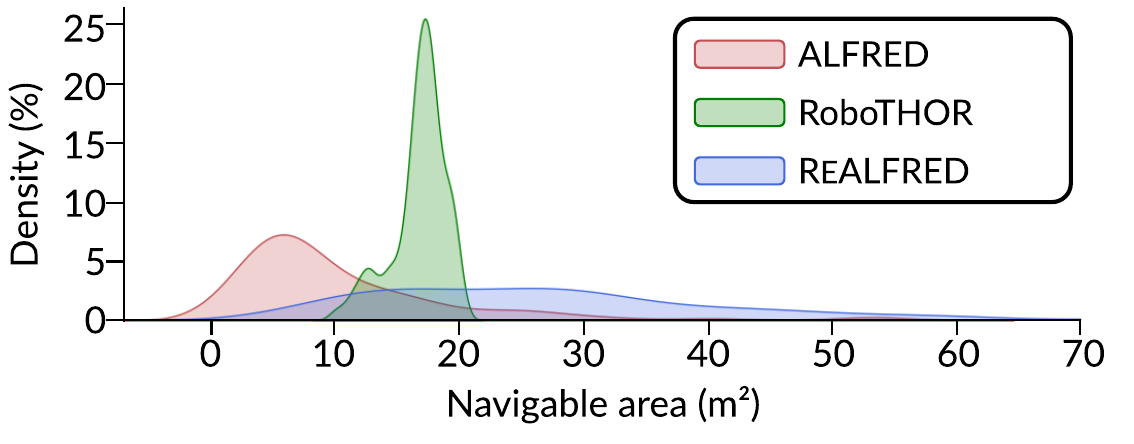}
        \caption{Distribution of navigable areas by scene.}
        \label{fig:nav_area}
    \end{subfigure}
    \caption{
        \textbf{Distribution of navigable and floor areas in interactive benchmarks by scenes.}
        `Floor area' denotes the overall size of the scene.
        `Navigable area' denotes the size of the space in which the agent can actually navigate.
        For both metrics, the \method benchmark poses a more even distribution and provides larger areas.
        For (a), we exclude RoboTHOR since it consists of single-sized floors.
    }
    \label{fig:distribution_area}
\end{figure}

\noindent \textit{Spatial size.}
We compare the \method benchmark with other benchmarks in terms of spatial size~\cite{ramakrishnan2021hm3d} by measuring `Floor area' and `Navigable area' and provide the result in Fig.~\ref{fig:distribution_area}.
`Floor area' represents the total spatial size ($m^2$) of a scene, defined by the floor projection.
`Navigable area' measures the spatial size ($m^2$) of the space in which an agent can actually navigate.
`Navigable area' is smaller than or equal to `Floor area' since `Navigable area' excludes areas where the agent collides with any components in the scene from `Floor area.'

We observe that the \method benchmark yields a diverse distribution of floor areas, compared to previous work~\cite{shridhar2020alfred}, with a larger average per scene. 
Furthermore, the \method benchmark provides a broader range of navigable areas, with larger areas on average.
This implies that an agent needs the ability to navigate effectively in spaces of varying sizes, generally wider on average.
\begin{table}[t!]
    \caption{
        \textbf{Component sizes of interactive embodied AI benchmarks.}
        Each `Total floor area' and `Total navigable area' (Total nav. area) denotes the sum of all floor areas and navigation areas.
        `Nav. complex.' denotes navigational complexity.
        `Scene clutter' measures the amount of clutter in the scene.
        We do not compare ReplicaCAD's Navigable area, Navigation complexity, and Scene clutter as agent sizes differ across simulators.
        The highest value for each metric is shown in \textbf{bold}.
        $^\dagger$We count objects used as target objects in the ObjectNav task.
        $^\ddagger$We count objects used in any of the tasks.
    }
    \centering
    \resizebox{0.97\columnwidth}{!}{
        \begin{tabular}{lcccc}
        \toprule
        Simulator           & Habitat2.0    & \multicolumn{3}{c}{AI2THOR} \\
        \cmidrule(lr){1-1} \cmidrule(lr){2-2} \cmidrule(lr){3-5}
        Dataset             &ReplicaCAD\cite{szot2021habitat}    & RoboTHOR\cite{deitke2020robothor}      & ALFRED\cite{shridhar2020alfred}      & \textbf{\method} \\
        \cmidrule(lr){1-1} \cmidrule(lr){2-2} \cmidrule(lr){3-5}
        \# Scenes          &$111$        & $75$           & $120$         & $\mathbf{150}$                 \\
        Total floor area ($m^2$)    &$8,824.5$     & $2,574$        & $2,555$         & $\mathbf{10,060}$                 \\
        Total nav. area ($m^2$)&$-$          & $1,258$        & $1,356$         & $\mathbf{4,251}$                   \\
        Nav. complex.         &$-$          & $2.036$        & $2.549$         & $\mathbf{3.020}$                   \\
        Scene clutter         &$-$          & $\mathbf{8.095}$        & $5.119$         & $8.072$                   \\
        \# object class      &$92$         & $14$$^\dagger$ & $82$$^\ddagger$  & $\mathbf{112}$$^\ddagger$                   \\
        \bottomrule
        
        \end{tabular}  
    }
    \label{tab:scale_comparison}
\end{table}

We also investigate the spatial characteristics of each benchmark~\cite{szot2021habitat, deitke2020robothor, shridhar2020alfred} and summarize the result in Table~\ref{tab:scale_comparison}.
`Total floor area' denotes the sum of all floor areas and `Total navigable area' denotes the sum of all navigable areas in the dataset.
We observe that the \method benchmark has the highest total floor area and navigable area value, implying that \method provides more navigation space for an agent than previous work~\cite{deitke2020robothor, shridhar2020alfred, szot2021habitat}.

\noindent \textit{Spatial complexity.}
We now investigate the complexity of spatial structures in the \method benchmark using several metrics.
To ensure a fair comparison with other datasets~\cite{deitke2020robothor, shridhar2020alfred, szot2021habitat}, we employ the navigation complexity introduced by \cite{xia2018gibson} and the scene clutter measurement from \cite{ramakrishnan2021hm3d}.
A higher navigation complexity indicates an increased difficulty in navigating through the space, while a higher scene clutter implies the presence of more obstacles in the environment.
By utilizing these metrics, we conduct a comparative analysis with object-interactable benchmarks~\cite{deitke2020robothor,shridhar2020alfred} and provide the result in Table~\ref{tab:scale_comparison}.

We observe that the \method benchmark provides the environment with higher navigation complexity and scene clutter compared to other benchmarks~\cite{shridhar2020alfred, szot2021habitat}.
The high navigation complexity in our scenes stems from their multi-room composition.
This setup requires the agent to execute more intricate navigation when moving from one room to another, in comparison to scenarios where the agent operates within a single room.
We observe that ours poses the second-highest scene clutter with a marginal gap from \cite{deitke2020robothor}.
This is because the spaces in \cite{deitke2020robothor} are relatively confined with a high density of furniture.
We observe that ours has a similar value to \cite{deitke2020robothor}, meaning that our scenes have a large amount of obstacles with a similar portion of \cite{deitke2020robothor} since our scenes' average size is larger than that of \cite{deitke2020robothor}.
This implies that the \method benchmark provides a complex and challenging space for the agent to explore the environment.

\subsection{Expert Demonstration Generation}
Each expert demonstration includes a set of an egocentric RGB view and action information with an interaction mask if exists at each time step.
Expert demonstrations for each task are generated by a planner~\cite{hoffmann2001ff} with encoded state spaces into Planning Domain Definition Language (PDDL) rules~\cite{aeronautiques1998pddl}.
To generate household tasks, we utilize seven task types introduced in \cite{shridhar2020alfred}.

\begin{figure}[t!]
    \centering
    \includegraphics[width=.95\linewidth]{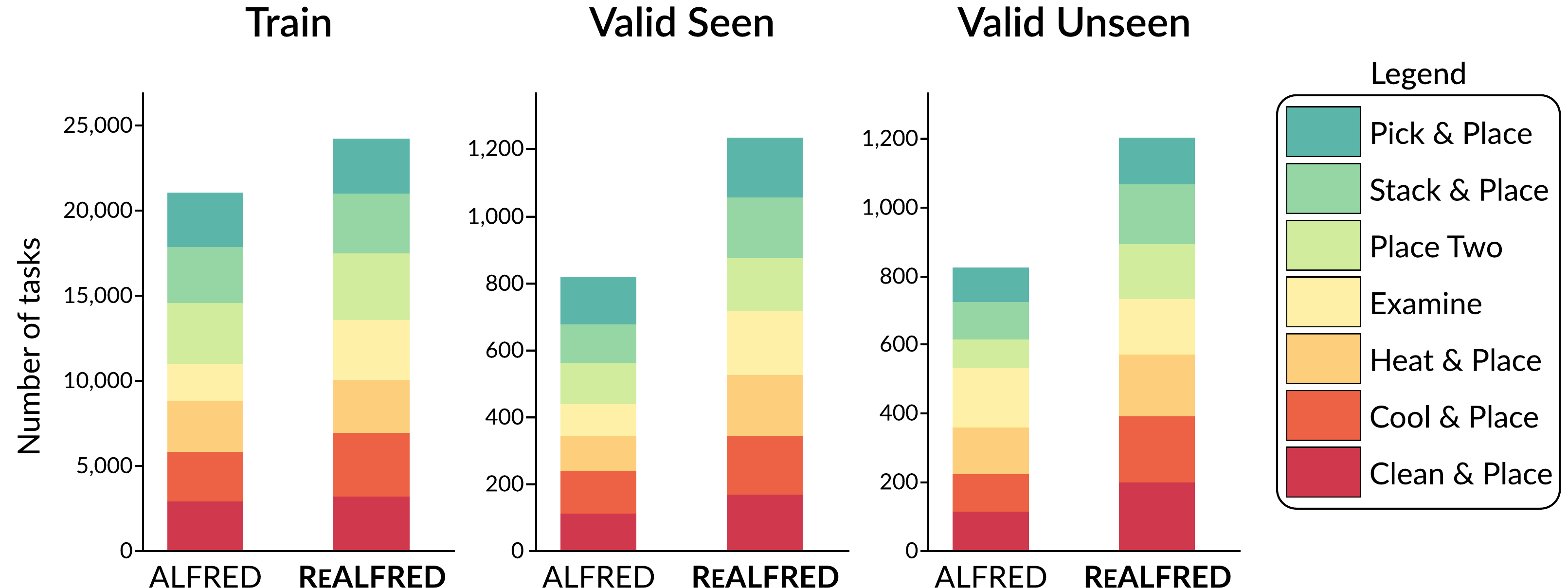}
    \caption{\textbf{The seven types of tasks' distribution in \method.}
    We provide $37.6\%$ more tasks in valid sets and $19.3\%$ in total compared to previous benchmark~\cite{shridhar2020alfred}.}
    \label{fig:task_type}
\end{figure}

\begin{wrapfigure}{r}{.45\textwidth}
    \centering
    \includegraphics[width=\linewidth]{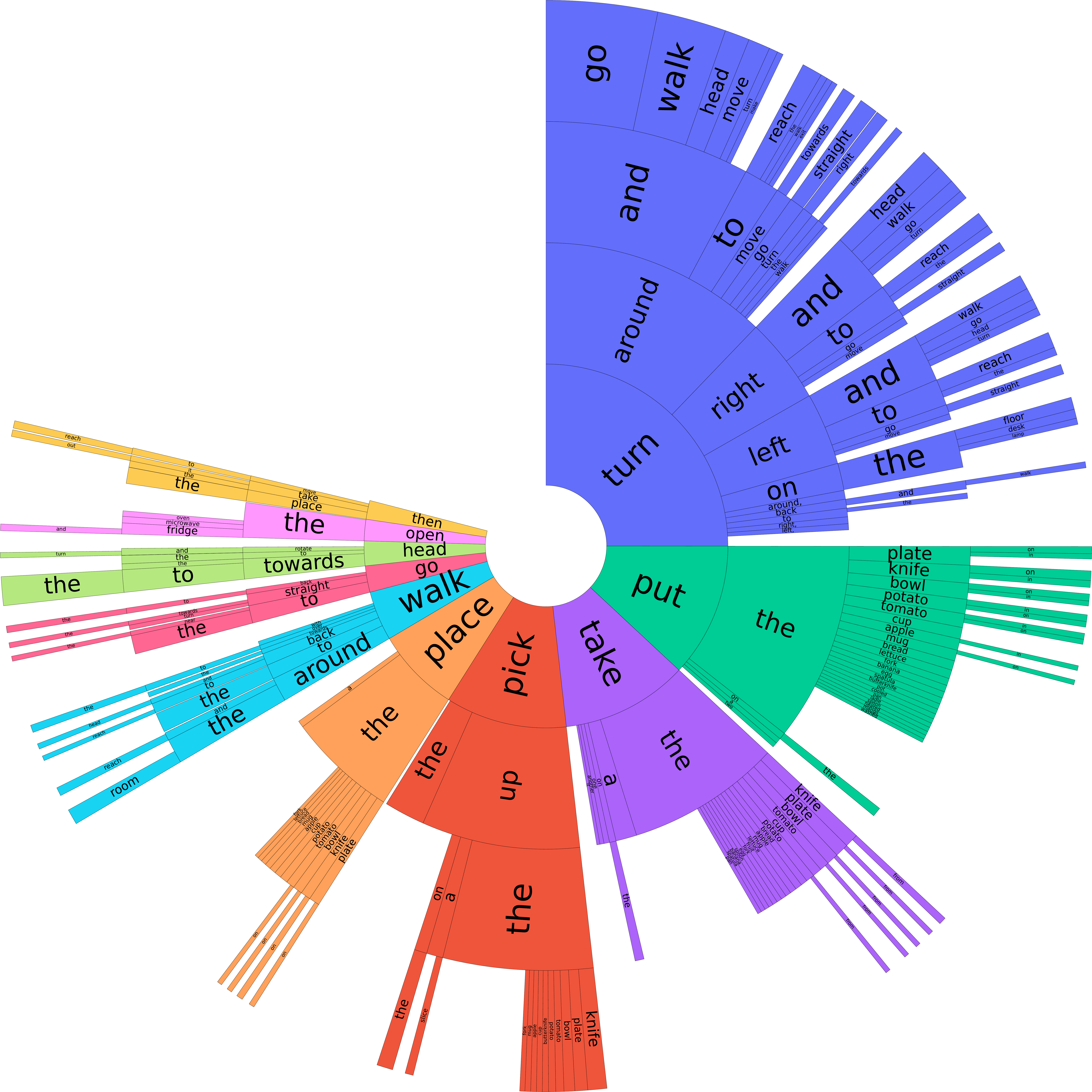}
    \caption{
        \textbf{Word distribution.}
        The words are arranged starting from the center and extending outwards.
        The arc length corresponds to the word frequency in the instructions.
    }
     \label{fig:language_dist}
\end{wrapfigure}
\noindent \textbf{Data splits.}
We split the generated demonstrations into train, validation, and test folds.
Specifically, we designate $135$ scenes for \textit{seen} and $15$ scenes for \textit{unseen} fold.
Comparison of the amount of each task and the distribution across training and validation folds with previous work~\cite{shridhar2020alfred} is shown in Fig.~\ref{fig:task_type}.

\noindent \textbf{Curating free-form language instructions.}
Detailed language instructions describe a task that involves a sequence of actions, a point of interest in robotics.
The \method benchmark offers $30,696$ language directives, each comprising a human-annotated high-level goal and a set of step-by-step instructions.
These directives are collected from $93$ Amazon Mechanical Turk workers with a `Master' qualification, ensuring high-quality.
Collected annotations are validated through an additional voting survey, and invalid instructions are replaced with newly collected instructions.
The distribution of language instructions by their first four words is presented in Fig.~\ref{fig:language_dist}.
We provide a detailed annotation process and examples of an expert demonstration with the instruction in Sec. A.1 and A.4 in the supplementary material.

\begin{wrapfigure}{r}{.45\textwidth}
\centering
\includegraphics[width=\linewidth]{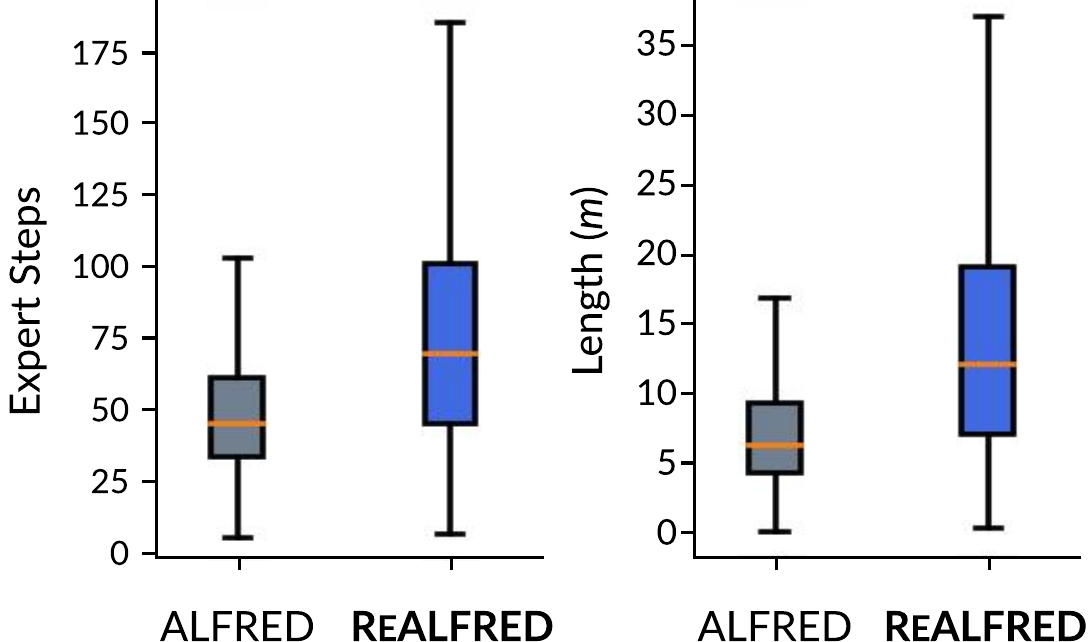}
\caption{\textbf{Comparison of expert demonstrations (ALFRED \textit{vs.} \method).}
Due to expanded environments, longer steps and planning horizons are required in \method.}
\label{fig:step_comparison}
\end{wrapfigure}
\noindent \textbf{Multi-room embodied instruction following.}
For household robots to be able to effectively assist humans, it would be more practical to deploy them in complex and diverse environments with various room types, rather than confining them to a single room.
Cross-room navigation poses significant challenges, requiring agents to understand room references, plan efficient paths, and overcome obstacles proficiently.
This agent is further tasked to interpret visual cues, demonstrate a keen sense of spatial awareness, and process natural language instructions.

The distribution of the number of steps and the length covered by the expert demonstrations is shown in Fig.~\ref{fig:step_comparison}.
We observe that longer steps and trajectories are required to complete our tasks compared to the single-room constraint benchmark~\cite{shridhar2020alfred}, meaning that our benchmark provides longer-horizon tasks.

To broaden a range of tasks, we provide an extensive number of object class types in the \method benchmark.
More object classes are added to the \method benchmark, making it a superset of \cite{shridhar2020alfred} with $86$ pickupable and $26$ receptacle objects.
This has resulted in a more diverse set of tasks, with the number of unique tasks being $84.3\%$ more than the ALFRED benchmark~\cite{shridhar2020alfred}, and the proportion of unique tasks being higher in the \method benchmark.
We provide detailed information in Sec.~\ref{sec:supp_bechmarkDetails} in the supplementary material.


\section{Experiments}
\noindent \textbf{Metrics.}
We follow the same evaluation protocol of the ALFRED benchmark~\cite{shridhar2020alfred}.
The primary metric is `Success Rate (SR)' which measures the percentage of completed tasks.
`Goal-Condition Success Rate (GC)' measures the percentage of achieved goal conditions.
We also use path-length-weighted metrics to measure how efficiently an agent completes tasks.
For more details, kindly refer to \cite{shridhar2020alfred}.

\noindent \textbf{Baselines.}
We evaluate several recent state-of-the-art methods~\cite{shridhar2020alfred, singh2021factorizing, kim2021agent, blukis2021persistent, min2021film, song2023llmplanner, kim2023context} with competitive results in \cite{shridhar2020alfred}.
We provide more details in Sec. B.

\newcommand{\mcc}[1]{\multicolumn{#1}{c}}
\definecolor{Gray}{gray}{0.90}
\newcolumntype{a}{>{\columncolor{Gray}}r}
\newcolumntype{b}{>{\columncolor{Gray}}c}
\newcommand{\B}[1]{\textcolor{blue}{\textbf{#1}}}

\begin{table*}[t!]
    \caption{
        \textbf{Task and Goal-Condition Success Rate.}
        We train and evaluate recent state-of-the-art methods on our \method benchmark.
        For a fair comparison, we group these methods into two based on the usage of extra depth supervision: models learned by imitation learning without depth supervision (`Imitation Learning') and ones that maintain semantic spatial representations constructed by predicted depth maps (`Spatial Map Reconst.').
        Path-length-weighted (PLW) metrics are reported in Sec. C for each value.
        $^\dagger$Authors' implementation as the code is not publicly available.
    }
    \centering
    \resizebox{1.00\textwidth}{!}{
        \begin{tabular}{@{}llaarraarr@{}}
            \toprule
            \multirow{3}{*}{Learning}& \multirow{3}{*}{Model}
                             & \mcc{4}{\textbf{Validation}} & \mcc{4}{\textbf{Test}} \\
                             \cmidrule{3-10}
                             & & \mcc{2}{\textit{Seen}} & \mcc{2}{\textit{Unseen}}
                             & \mcc{2}{\textit{Seen}} & \mcc{2}{\textit{Unseen}}  \\
                             & & \multicolumn{1}{b}{SR} & \multicolumn{1}{b}{GC} 
                             & \multicolumn{1}{c}{SR} & \multicolumn{1}{c}{GC}
                             & \multicolumn{1}{b}{SR} & \multicolumn{1}{b}{GC} 
                             & \multicolumn{1}{c}{SR} & \multicolumn{1}{c}{GC} \\
            
            \cmidrule{1-10}
            
            \multirow{3}{*}{\parbox[c]{2cm}{\centering\textbf{Imitation \\Learning}}} & {Seq2Seq~\cite{shridhar2020alfred}}     & $0.77\pm0.06$     & $6.93\pm0.06$     & $0.00\pm0.00$    & $4.03\pm0.00$   & $1.10\pm0.00$    & $6.60\pm0.00$   & $0.00\pm0.00$  & $3.50\pm0.00$  \\
            & {MOCA~\cite{singh2021factorizing}}      & $12.64\pm0.12$      & $20.95\pm0.18$     & $1.44\pm0.05$      & $6.76\pm0.04$     & $14.11\pm0.03$     & $22.84\pm0.04$   & $0.62\pm0.08$    & $5.14\pm0.08$  \\
            & {ABP$^\dagger$~\cite{kim2021agent}}               & $24.71\pm0.05$     & $33.80\pm0.14$     & $4.22\pm0.05$      & $11.71\pm0.27$    & $27.44\pm0.40$    & $35.81\pm0.23$    & $3.54\pm0.23$  & $10.57\pm0.22$  \\
            \cmidrule{1-10}
            
            \multirow{4}{*}{\parbox[c]{2cm}{\centering\textbf{Spatial\\Map\\Reconst.}}} & {HLSM~\cite{blukis2021persistent}}      & $4.23\pm0.08$             & $9.14\pm0.09$             & $1.08\pm0.14$            & $6.12\pm0.23$          & $6.27\pm0.04$              & $10.44\pm0.13$           & $0.49\pm0.16$             & $4.28\pm0.13$  \\
            & {FILM~\cite{min2021film}}                & $7.08\pm0.28$       & $11.93\pm0.23$       & $4.44\pm0.17$      & $9.26\pm0.13$    & $8.79\pm0.07$     & $13.03\pm0.08$       & $2.15\pm0.18$       & $6.56\pm0.15$  \\
            & {LLM-Planner$^\dagger$~\cite{song2023llmplanner}}   & $5.80\pm0.19$       & $11.69\pm0.35$       & $3.33\pm0.22$      & $8.29\pm0.19$    & $8.16\pm0.20$     & $13.20\pm0.13$       & $1.90\pm0.13$       & $6.33\pm0.02$  \\
            & {CAPEAM$^\dagger$~\cite{kim2023context}}           & $13.45\pm0.05$       & $18.16\pm0.27$       & $4.92\pm0.22$      & $9.47\pm0.23$    & $15.61\pm0.15$     & $20.22\pm0.11$       & $2.87\pm0.13$       & $7.36\pm0.07$  \\
            
            \cmidrule{1-10}
            
            & {Human}     & \multicolumn{1}{b}{-} & \multicolumn{1}{b}{-} & \multicolumn{1}{c}{-} & \multicolumn{1}{c}{-} & \multicolumn{1}{b}{-} & \multicolumn{1}{b}{-} & $85.00\pm3.54$  & $91.30\pm2.94$ \\

            \bottomrule
        \end{tabular}
    }    
    \label{tab:baseline-result}
\end{table*}

\subsection{Comparison of the State of the Arts}
We evaluate the baselines in the proposed \method benchmark over multiple runs and present the average result in Table~\ref{tab:baseline-result}.
We report extended results with path-length-weighted metrics in Sec.~\ref{sec:supp_extended_qauntitative} in supplementary.

For a fair comparison, we separate the baseline into two groups based on the use of additional depth supervision for semantic map reconstruction: `Imitation Learning' where agents learn direct mapping from visual observations and language instructions to action sequences and `Spatial Map Reconst.' where agents plan action sequences based on reconstructed semantic spatial representations.

We observe that all these baselines, `Imitation Learning' and `Spatial Map Reconst.,' consistently achieve lower performance values for all metrics in both \emph{seen} and \emph{unseen} splits compared to performances achieved in \cite{shridhar2020alfred}, implying that our proposed \method provide more challenges compared to \cite{shridhar2020alfred}.
While in \cite{shridhar2020alfred}, the `Spatial Map Reconst.' baselines~\cite{blukis2021persistent,min2021film,kim2023context} outperform learning-based approaches~\cite{shridhar2020alfred,singh2021factorizing,kim2021agent} by exploiting semantic spatial representation with deterministic algorithms (\eg, obstacle-free navigation path planning~\cite{sethian1996fast}), we observe a contrasting result that the `Imitation Learning' baselines outperforming the `Spatial Map Reconst.' baselines in our \method benchmark.

We qualitatively observe that such a confined spatial map reconstruction is due to a limited field of view and map reconstruction methods.
The agent's limited field of view often leads to failure in recognizing room corners with doors or narrow aisles, resulting in (single-room-sized) limited map reconstruction.
In addition, \cite{min2021film, kim2023context} perceives obstacles larger than they actually are for better obstacle-free path planning by sacrificing navigable area, but this can be quite critical for narrow doors and aisles as they have a small amount of navigable space.
This may hinder navigation to other rooms and thus, fail at tasks.
We provide a more detailed discussion, supported by figures, on the agents' difficulty in recognizing narrow passages and the resulting impact on the agents, who struggle with spatial map reconstruction in Sec.~\ref{sec:supp_map_recon} in the supplementary material.

\subsection{Comparison to the agents with sim-to-real adaptation}
We investigate the transfer from simulation training to real-world scan evaluation (sim-to-real) and from real-world scan training to real-world scan evaluation (real-to-real) with \cite{kim2021agent}.
We train a sim-to-real agent with synthetic visual data and a real-to-real agent with real scanned visual data, respectively.
During inference, both agents predict an action and an object mask based on the scanned visual input frame and the given language instruction at every time step.

\noindent \textbf{Training and evaluation data selection.}
For training the sim-to-real agent, we use the training dataset in \cite{shridhar2020alfred}, encompassing $21K$ language annotations.
For a fair comparison, we train the real-to-real agent with tasks from the \method benchmark's training fold, specifically involving the manipulation of the same objects that were used in training the sim-to-real agent with $19K$ language annotations.
For evaluation, we select tasks from the valid unseen splits of our \method benchmark.
We evaluate the agent performance with the tasks that 1) do not require multi-room navigation 2) and those that do.
This selection specifically includes tasks that feature objects used in the training phase.

\noindent \textbf{Real-to-sim domain adaptation.}
The use of Generative Adversarial Networks (GANs)~\cite{goodfellow2014generative, CycleGAN2017} for domain adaptation has recently been examined in the literature on robotics~\cite{rao2020rl,bousmalis2018using, zhang2019vr, ho2021retinagan, khansari2023practical}.
These models are employed to adapt input images from real-world domains (\ie Real $\rightarrow$ Sim) before they are passed to the agent policy.
We train a CycleGan~\cite{CycleGAN2017} and its off-the-shelf variant~\cite{torbunov2023uvcgan} with unpaired images collected from the \method (real domain) and ALFRED benchmarks~\cite{shridhar2020alfred} (simulated domain).
Among the trained generators, the one performing real to simulation conversion, referred to as the \textit{real-to-sim goggle}, learns the mapping from the real domain to the simulation domain $G_S : \mathcal{R} \rightarrow \mathcal{S}$, where $\mathcal{R}$ denotes the real domain and $\mathcal{S}$ denotes the simulated domain.

\begin{wraptable}{r}{.45\textwidth}
    \centering
        \caption{\textbf{Comparison to the agents with sim-to-real adaptation.}
    Each `Sim2Real' and `Real2Real' denotes an agent trained in simulated environments and 3D-captured scenes.
    `Goggle' denotes \emph{real-to-sim} methods.
    }
    \resizebox{\linewidth}{!}{
    \begin{tabular}{@{}cllbbcc@{}}
        \toprule
         & & & \multicolumn{2}{c}{\textbf{Multi + Single}}&  \multicolumn{2}{c}{\textbf{Single only}} \\
         \cmidrule(lr){4-5} \cmidrule(lr){6-7}
        \textbf{\#}    & \textbf{Setting}             & \textbf{Goggle}                                    & \textbf{SR}                    & \textbf{PLWSR} & \textbf{SR}                    & \textbf{PLWSR}    \\
        \midrule
        ($a$)     & Sim2Real   & \textit{None}                                       & $0.115$   & $0.012$ & $0.0$   & $0.0$   \\
        ($b$)     & Sim2Real    & CycleGan~\cite{CycleGAN2017}              & $0.115$   & $0.016$ & $0.327$   & $0.065$   \\
        ($c$)     & Sim2Real    & UVCGAN-v2~\cite{torbunov2023uvcgan}       & $0.115$   & $0.046$ & $0.327$   & $0.187$   \\
        \midrule
        ($d$)     & Real2Real           & \textit{None}                                      & $2.405$   & $0.785$  & $2.614$   & $0.762$   \\
        \bottomrule
        \label{tab:sim2real}
    \end{tabular}%
    }
\end{wraptable}
\noindent \textbf{Results.}
We present the results of the sim-to-real experiments in Table~\ref{tab:sim2real}.
We report the performance of agents when tasks 1) require agents to navigate multi-rooms, denoted as `Multi + Single,' and 2) are solvable within a single room, denoted as `Single only.'
Firstly, we observe that the sim-to-real agent significantly underperforms its real-to-real counterpart in all metrics ($\#(a)$ \vs $\#(d)$).
We then compare the results with \textit{goggled} agents.
By comparing agents evaluated in the `Multi + Single' tasks, we do not observe improvements in the main metric, success rate (SR).
We observe a slight increase in the PLWSR metric for the \textit{goggled} sim-to-real agent compared to the vanilla sim-to-real agent ($\#(a)$ \vs $\#(b, c)$).

To isolate the impacts of the visual domain gap, we then compare agents evaluated on `Single only' tasks.
We observe that both \textit{goggled} agents show improvements over the vanilla sim-to-real agent which results in zero success rate.
However, a noticeable performance gap exists with the real-to-real agent, implying the need for learning in real scanned environments ($\#(d)$ \vs $\#(a, b, c)$).

\subsection{Challenges in \method}
We propose two hypotheses for the low performance observed with state-of-the-art methods on the \method benchmark: (1) navigating within larger scenes and (2) overcoming narrow pathways between rooms.
These elements represent challenges of completing tasks within multi-room, household-scale environments. 
We analyze failure cases with the best-performing model, ABP~\cite{kim2021agent}, on \textit{valid unseen} split, since it seems the most promising on the \method benchmark.

\begin{figure*}[t!]
    \centering
    \includegraphics[width=.975\linewidth]{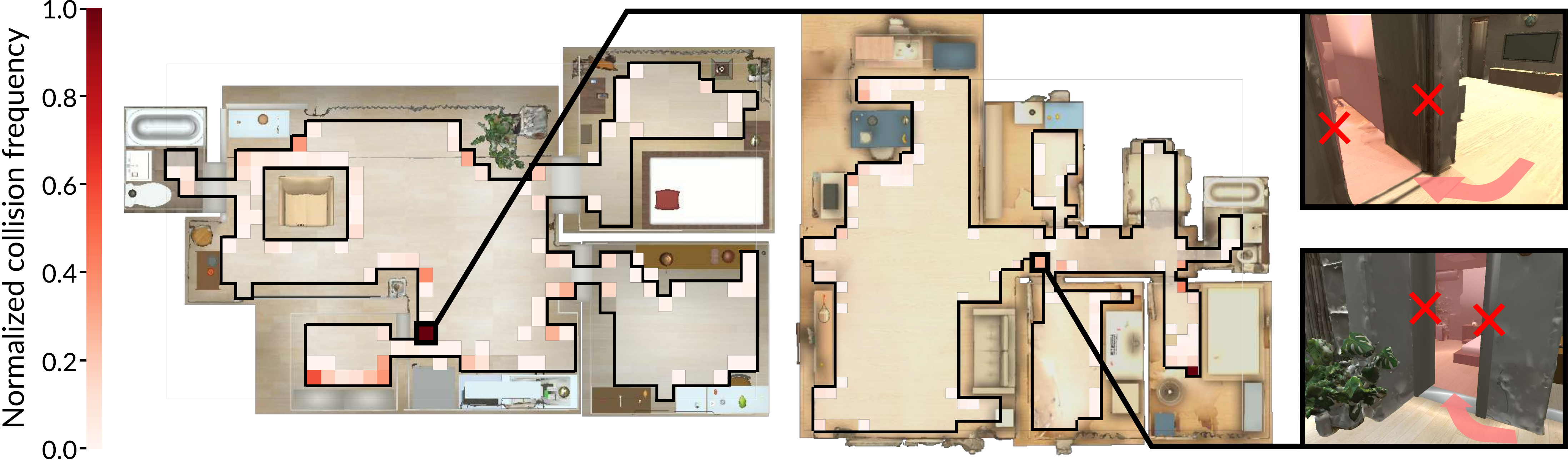}
    \caption{\textbf{Distribution of collision spots.}
    The heatmap illustrates collision frequency on each spot and the black outline depicts a navigable area.
    We illustrate collision points of \cite{kim2021agent} in \textit{valid unseen} environments to represent the distribution.
    Specifically, we choose failure episodes from the two largest scenes in the \textit{valid unseen} split.
    We observe a remarkable concentration of collisions at the entrance to the rooms.
    The corresponding egocentric views are presented on the right side of the figure.
    }
    \label{fig:failure_quali}
\end{figure*}

\noindent \textbf{Difficulty in large environments.}
Understanding the surrounding environment, including the location of objects and the positions of obstacles, can be beneficial for task completion.
However, the agent's visual range is bounded, requiring it to expend more steps for exploration in order to perceive larger spaces.
This intensifies the challenge of navigating larger spaces with limited steps.

We conduct an analysis to see the relationship between the success rate (SR) and the size of the space.
We follow the interquartile range (IQR) method to set a threshold at $30.44 m^2$ that covers the most navigable space sizes in the ALFRED~\cite{shridhar2020alfred}.
This threshold marks the upper fence, defining spaces above it as outliers.
We classify spaces in the \method benchmark smaller than threshold as smaller scenes, and those larger as larger scenes.
Results indicate that the agent~\cite{kim2021agent} with the highest SR in the \textit{valid unseen} fold showed an average SR of $5.46\%$ in the smaller scenes and only $1.77\%$ in the larger scenes.
This implies that solving tasks becomes more challenging as the space size increases.

In addition, we compare the difficulty of navigation between the previous~\cite{shridhar2020alfred} and \method benchmarks with different average spatial sizes.
To quantify, we first define milestones for each task, which are spots to be reached to interact with target objects (\eg, Apple, Knife, \etc).
We consider navigation to be a success only if \emph{all} milestones are visited, regardless of whether the actual interaction (\eg, slicing an apple) is performed or not.
Consequently, the navigation success rate in the \method benchmark is merely $59.18\%$ for the \textit{valid unseen} split while the agent's~\cite{kim2021agent} navigation success rate is $84.82\%$ in ALFRED~\cite{shridhar2020alfred}.
The \method benchmark offering an average space size more than three times larger than the previous work~\cite{shridhar2020alfred} implies the need for model development capable of overcoming the challenges associated with this increased scale.

\noindent \textbf{Navigation through narrow doorways.}
The \method benchmark supports environments with multiroom composition, unlike the previous dataset~\cite{shridhar2020alfred} providing a single room scale.
Specifically, it contains narrow doorways across rooms, which may hinder the agent from navigating through.
Here, we hypothesize that the agent would frequently collide with the walls near the door.

We investigate collision spots and showcase two examples on top of the scene's layout in \textit{valid unseen} fold in Fig.~\ref{fig:failure_quali}.
The number of collisions in failed cases is accumulated and normalized to the maximum number of collisions, respectively, to indicate the collision frequency at each point.
The black outline represents the agent's navigable area, as detailed in Sec.~\ref{sec:comparision_with_prior_benchmark}.
Each value denotes the normalized collision frequency.
In both scenes, we observe that collisions often occur on walls near the door, implying that our spatial characteristic (\ie, including narrow spaces) may hinder the agent from properly navigating without collision.

\noindent \textbf{Human evaluation.}
Following \cite{shridhar2020alfred}, we randomly select $100$ directives from the \textit{test unseen} fold and have them evaluated by humans.
Five participants are given $20$ tasks each, which they complete using a keyboard-and-mouse interface.
Before starting, they are given the opportunity to become familiar with the interface.

Participants achieve a comparable high success rate of $85\%$, along with a goal-condition success rate of $91.30\%$, in average.
We agree that human performance may appear slightly lower compared to the results presented in previous work~\cite{shridhar2020alfred, zhu2023excalibur, ramrakhya2022habitat}.
This is partly because human participants encountered several difficulties when controlling the agent, particularly in avoiding collisions within narrow corridors.
Furthermore, navigating large spaces with a limited egocentric field of view introduced additional challenges, leading to task failures.


\section{Conclusion}
\label{others}
We present \method, a new dataset and benchmark for embodied instruction following task on 3D-captured environments.
We capture $150$ indoor houses in 3D with interactable objects to enable complex household tasks.
The reconstructed indoor scenes provide a larger spatial area and complex multi-room environments that are close to the real-world scenario and challenging for an agent to successfully complete a task.
Expert demonstrations are also provided along with free-form human-language instructions.

In our empirical evaluations, we show that state-of-the-art methods struggle in large multi-room environments, provide analyses of our newly proposed benchmark, and perform Sim2Real transfer experiments.
We have released our Embodied AI research data and code for reproducibility.
We expect that the \method benchmark will encourage further research on developing robotic agents that execute household tasks by language instructions in the real world.

\noindent \textbf{Limitation and future work.}
Although we support a large number of interactable objects, the types of tasks to be completed are rather limited, considering more complex real-world scenarios.
In addition, we currently address natural language in English but users may come from different regions with different languages.
We can think of two future research avenues as follows. 
(1) adding additional complicated types of task that require both hands to complete. 
(2) supporting a multi-lingual interface for users from different regions.
\section*{Acknowledgements}
This work was partly supported by the NRF grant (No.2022R1A2C400230012, 5\%) and IITP grants (No.RS-2022-II220077 (5\%), No.RS-2022-II220113 (5\%), No.RS-2022-II220959 (5\%), No.RS-2022-II220871 (15\%), No.RS-2020-II201361 (5\%, Yonsei AI), No.RS-2021-II211343 (5\%, SNU AI), No.RS-2021-II212068 (5\%, AI Innov. Hub), No.RS-2022-II220951(50\%)) funded by the Korea government(MSIT).

%
%
\bibliographystyle{splncs04}
\bibliography{main}

\clearpage
\appendix
\newcommand{\origref}[1]{\textcolor{ForestGreen}{#1}}
\def\thefootnote{*}\footnotetext{Equal contribution. $^\dagger$Corresponding author.}

\noindent\textbf{Note:} \origref{Green} denotes reference to the main paper.

\section{Benchmark Details}
\label{sec:supp_bechmarkDetails}
\begin{figure}[t!]
    \centering
    \includegraphics[width=0.8\linewidth]{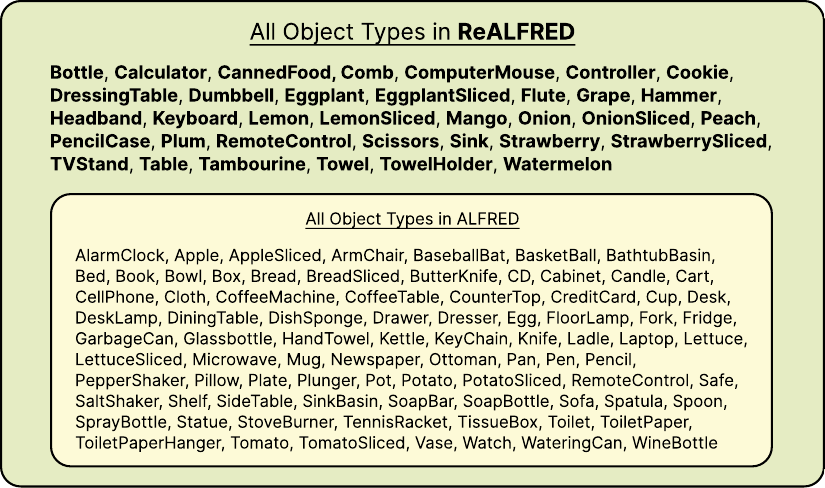}
    \caption{
        \textbf{Object list in ALFRED and \method.}
        Listed in alphabetical order.
        The object classes in the \method benchmark are a superset of those in ALFRED, with newly introduced objects highlighted in bold.
    }
    \vspace{-.5em}
    \label{fig:obj_list}
\end{figure}

We provide all 114 types of objects in Fig.~\ref{fig:obj_list}.
The \textbf{bold} text denotes the uniquely introduced object in ours.

\subsection{Annotation Interface}
\label{supp:annotation_interface}
\begin{figure}[t!]
\centering
    \begin{subfigure}{0.45\linewidth}
        \centering
        \includegraphics[width=0.9\linewidth]{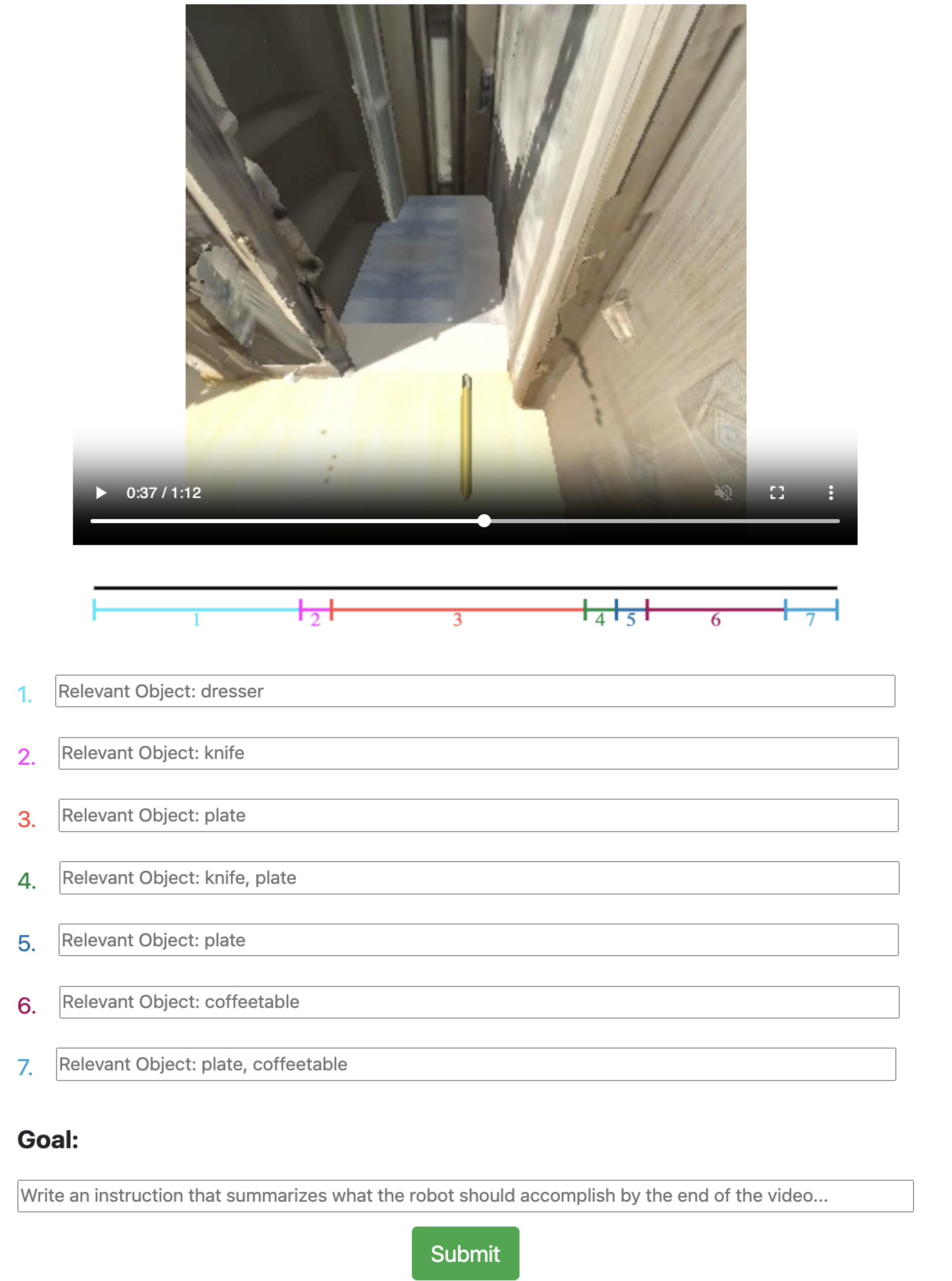}
        \caption{
        \textbf{Annotation interface.}
        }
        \label{fig:mturk_interface}
    \end{subfigure}
    \begin{subfigure}{0.45\linewidth}
        \centering
        \includegraphics[width=0.9\linewidth]{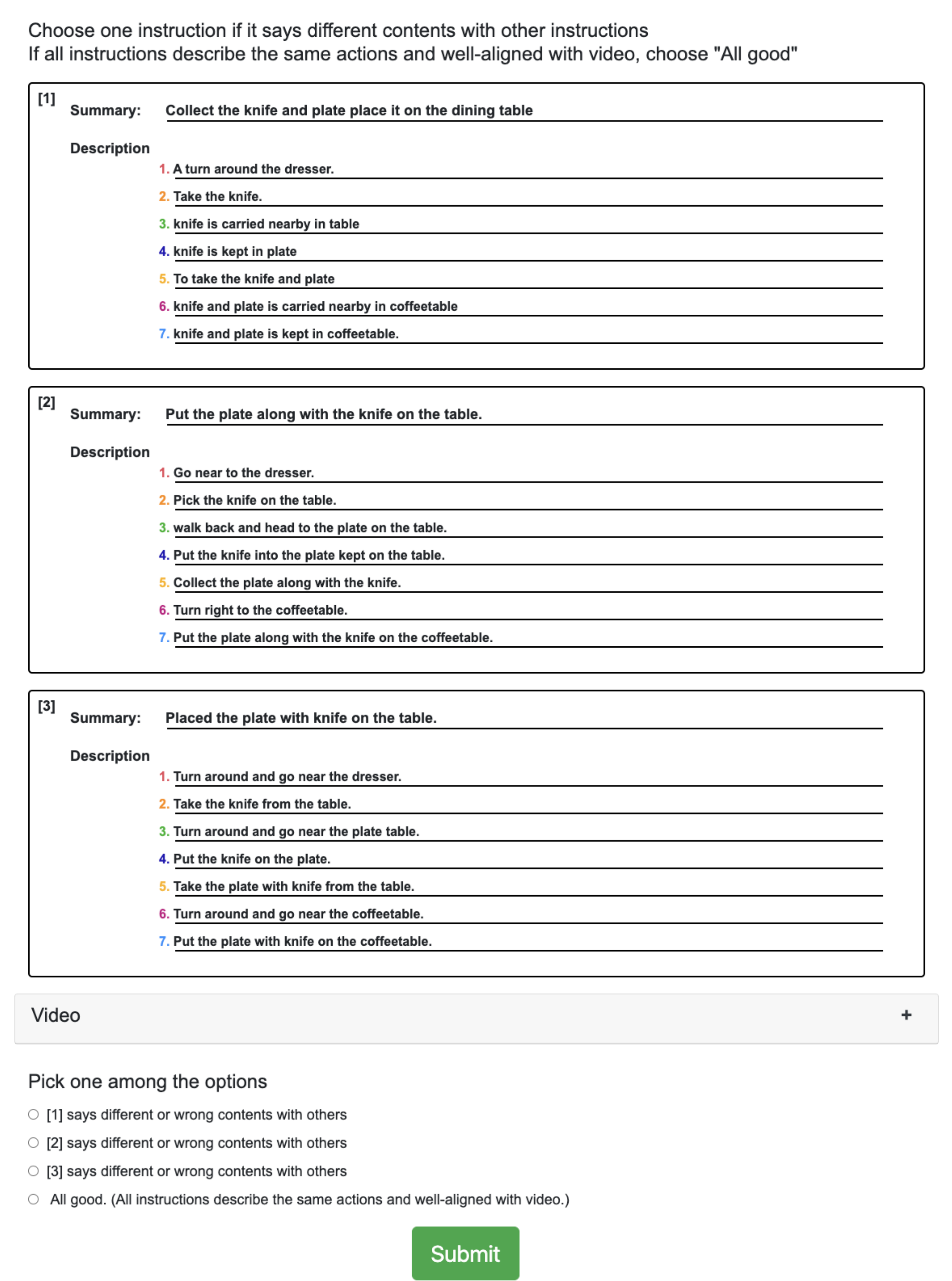}
        \caption{
        \textbf{Voting survey interface.}
        }
        \label{fig:mturk_voting}
    \end{subfigure}
    \caption{
        \textbf{Mechanical Turk interface.}
    }        
    \vspace{-1em}
\end{figure}

In this section, we describe the overall process of acquiring language annotations.
Fig.~\ref{fig:mturk_interface} illustrates the interface of the Mechanical Turk used to collect human annotations from Mechanical Turk workers.
We provided workers with an expert demonstration video and divided the timeline segments that have the intended subgoal (\eg, \textit{`pick up the pencil case,' `go to the sofa'}).
Workers were asked to fill each segment with their own words (\eg, \textit{`Pick up the pencil case from the coffee table,' `Walk around the table to get closer to the sofa'}).
Workers were paid \$0.7 per annotation as following the previous work~[\origref{61}].
Moreover, we adopted a voting survey to filter out inappropriate annotations.
Fig.~\ref{fig:mturk_voting} illustrates the interface of getting votes from workers.
We conducted the voting with a minimum of $2$ and up to $5$ reviewers per annotation.
Only annotations that received more than a majority of accepts in all cases were included in our set of annotations.
For annotations that did not achieve a majority of accepts, we re-collected annotations and implemented a voting system to prevent the inclusion of low-quality annotations.
We paid workers \$0.35 to compare $5$ sets of annotations following~[\origref{61}].

\subsection{Vocabulary Distribution}
\begin{figure*}[t!]
    \centering
    \includegraphics[width=0.98\linewidth]{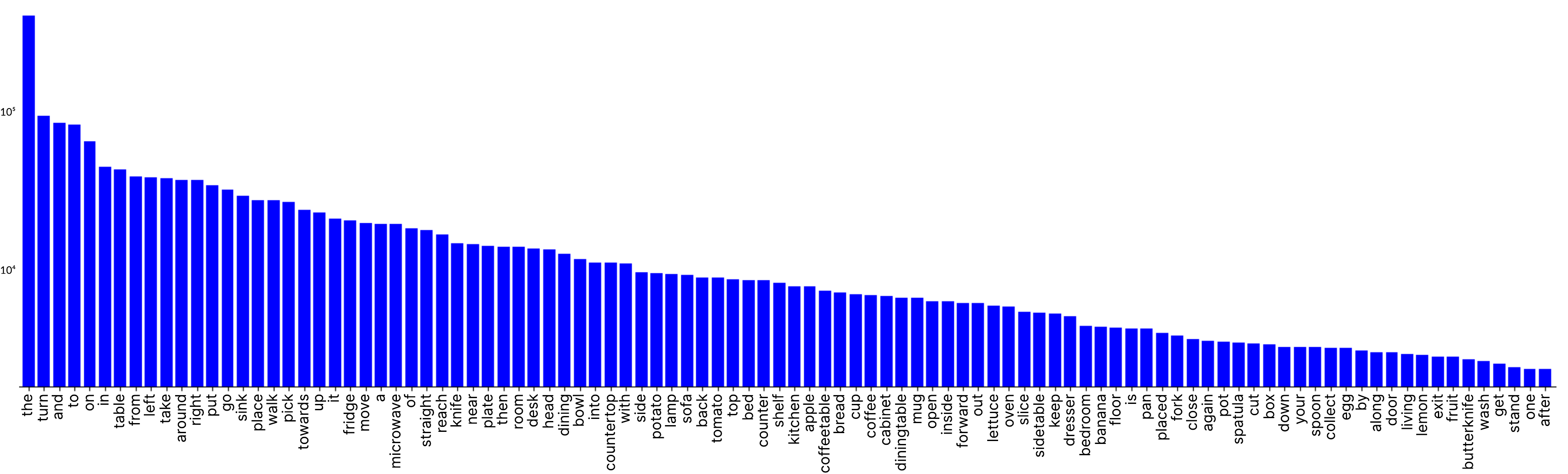}
    \caption{
        \textbf{Vocabulary statistics in collected human language instructions.}
    }
    \label{fig:supp_vocab}
\end{figure*}

We provide vocabulary statistics for the language instructions in the \method benchmark in Fig.~\ref{fig:supp_vocab}.

\subsection{Scanned Indoor Houses}
Among collected 150 scenes, we split scenes into 135 seen and 15 unseen environments.
Then we further split into validation (both seen and unseen) and test (both seen and unseen) folds.
The details are presented in Table~\ref{tab:scene_split}.
Note that validation and test unseen scenes are exclusive.

\vspace{-1.2em}
\subsubsection{Degree of photorealism.} We compare a degree of photorealism by measuring FID~[\origref{27}] and KID~[\origref{4}] scores following Ramakrishnan \etal~[\origref{56}].
We use rendederd RGB images from each synthetic environments~[\origref{20, 41}].
We compare the image quality with a set of RGB images rendered from previous dataset, HM3D~[\origref{56}] and Gibson~[\origref{73}].
To compare with previous scanned environments, we acquire a collection of real RGB images derived from high-resolution raw panoramas in Gibson. 
We designate this collection as Gibson HQ.
Results are presented in Table~\ref{tab:photorealism_comparison}.
We observe that ours achieves the lowest (\ie, best) FID and KID scores when compared to the previous environments~[\origref{20, 41}] that provide interaction with the environments including the objects.
However, despite these promising metrics, our results reflect lower photorealism compared to the previous scanned environments~[\origref{73, 56}] which do not provide interactable environments.
This may be due to manually added agent-interactable objects.

\vspace{-1.2em}
\subsubsection{Manual removal of 3D objects from background.} Designers use automated tools (\ie, Blender) to extract objects from the background meshes easily and set their properties (\eg, labels, colliders, interactability, \etc).

\begin{figure}[t!]
    \centering
    \includegraphics[width=0.75\linewidth]{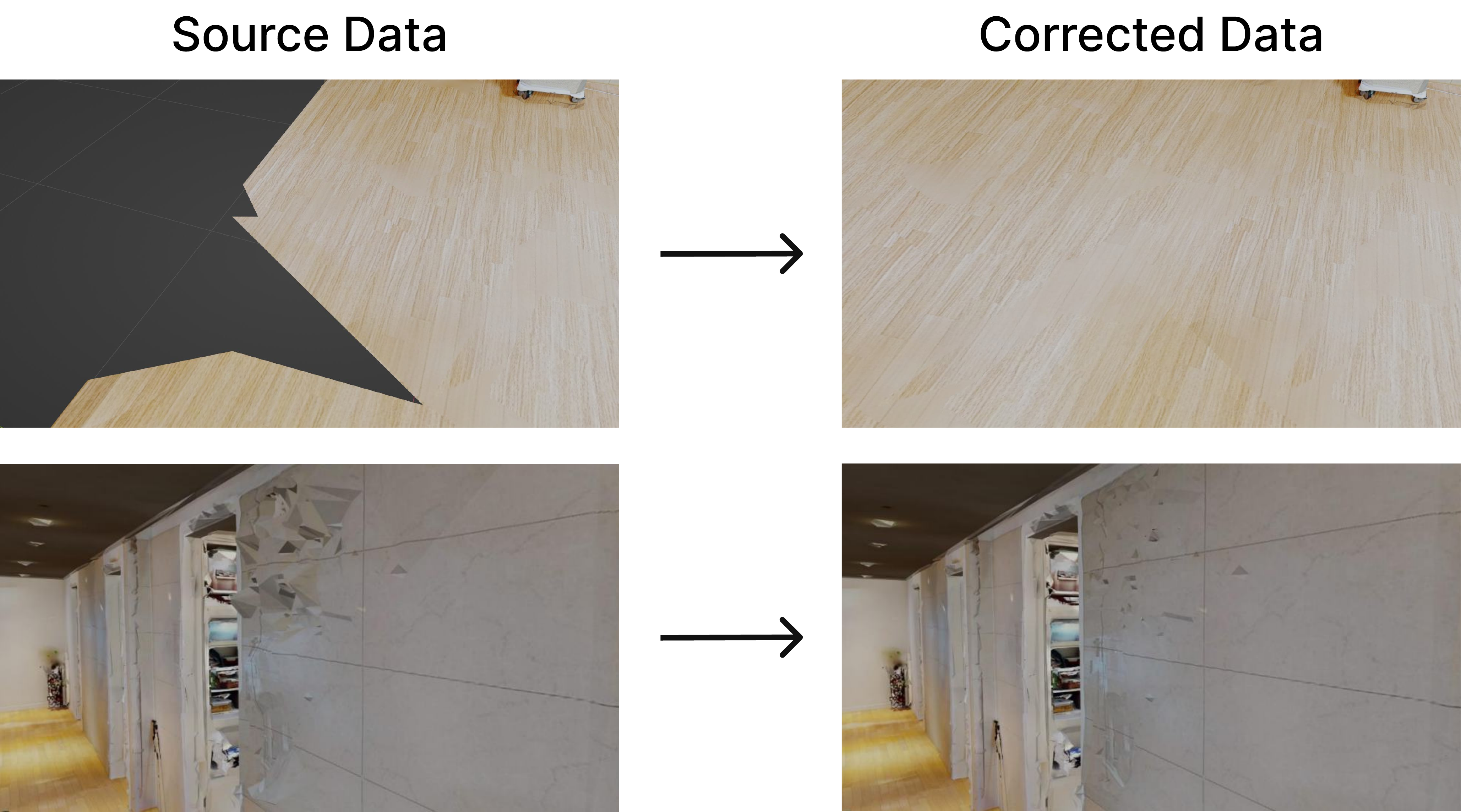}
    \caption{
        \textbf{Qualitative examples of the 3D mesh data correction.}
        In each row, we provide source data and the corrected data.
    }
    \label{fig:supp_data_correction}
\end{figure}

In Fig.~\ref{fig:supp_data_correction}, we provide the qualitative examples of how the source data were fixed by data correction.
Workers fill in the missing parts or smooth out the uneven surface by checking the overall alignment composition after mapping and alignment using automated tools such as Blender.

\vspace{-1.2em}
\subsubsection{Qualitative examples.} 
\begin{figure*}[t!]
    \centering
    \includegraphics[width=0.98\linewidth]{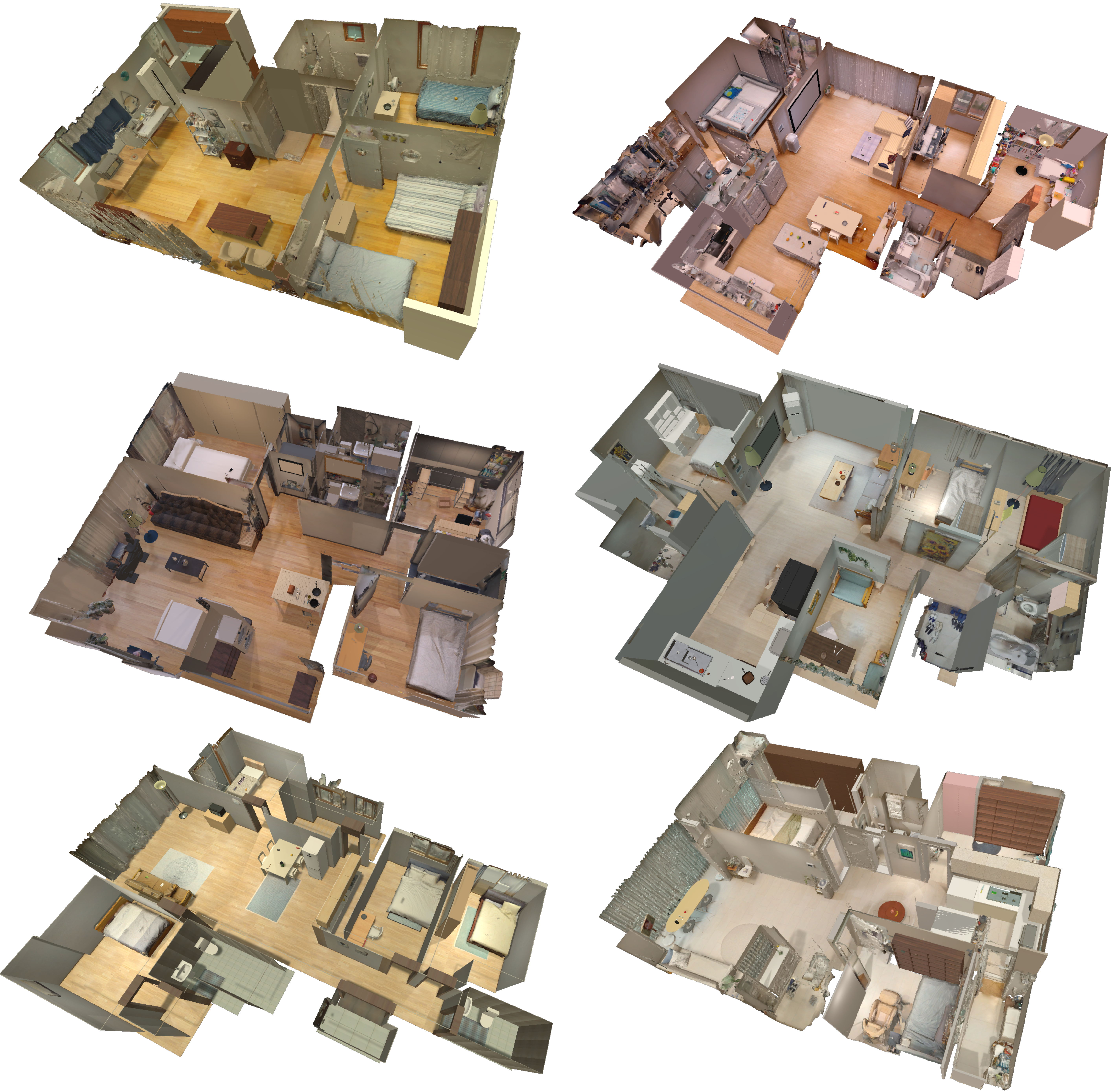}
    \caption{
        \textbf{Example of houses used in the \method benchmark.}
    }
    \label{fig:supp_best6}
\end{figure*}

\begin{figure*}[t!]
    \centering
    \includegraphics[width=0.98\linewidth]{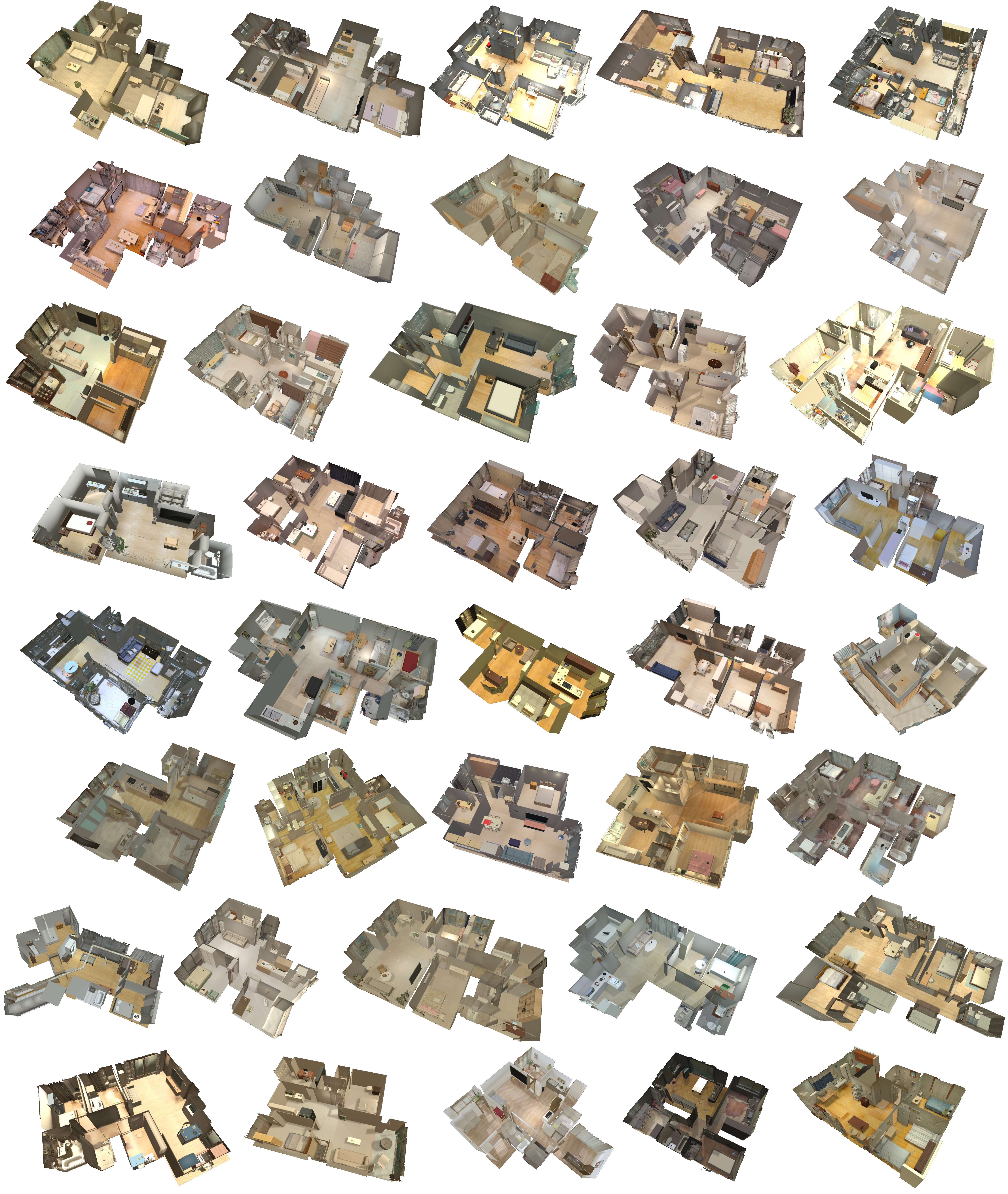}
    \caption{
        \textbf{Additional example of houses used in the \method benchmark.}
    }
    \label{fig:supp_rest_houses}
    \vspace{-1em}
\end{figure*}

We show several houses used in the \method benchmarks in Fig.~\ref{fig:supp_best6} and~\ref{fig:supp_rest_houses}.

\begin{table}[t!]
    \caption{
        \textbf{Indoor house splits.}
    }
    \centering
    \vspace{-0.5em}
        \begin{tabular}{lccccc}
        \toprule
                           & Train    & \multicolumn{2}{c}{Validation} & \multicolumn{2}{c}{Test} \\
        \cmidrule(lr){2-2} \cmidrule(lr){3-4} \cmidrule(lr){5-6}
                          &            & Seen   & Unseen                & Seen & Unseen \\
        \midrule
        \# of scenes       &   135       & 131    & 6                  & 135    & 9  \\
        \bottomrule
        \end{tabular}
    \label{tab:scene_split}
\end{table}

\begin{table}[t!]
    \caption{
        \textbf{Photorealism comparison.}
    }
    \centering
    \vspace{-0.5em}
    \resizebox{1.00\textwidth}{!}{
        \begin{tabular}{lcccccc}
        \toprule
            Environments           & \multicolumn{2}{c}{HM3D}                  & \multicolumn{2}{c}{Gibson}        & \multicolumn{2}{c}{Gibson HQ} \\
                    &    FID$\downarrow$ & KID$\times10^3\downarrow$      & FID$\downarrow$ & KID$\times10^3\downarrow$ & FID$\downarrow$ & KID$\times10^3\downarrow$\\
        \midrule
        TDW~[\origref{20}]         &   129.52           & 89.13$\pm$1.52       & 122.68          & 80.38$\pm$1.36  & 124.89    & 79.05$\pm$1.49 \\
        BEHAVIOR-1K~[\origref{41}] &   113.59           & 81.58$\pm$2.96       & 99.52           & 66.03$\pm$1.85  & 107.53    & 63.03$\pm$1.99 \\
        \textbf{\method}           &   \textbf{81.06}   & \textbf{69.96$\pm$1.47}       & \textbf{83.25}           & \textbf{71.05$\pm$1.51}  & \textbf{101.60}    & \textbf{89.98$\pm$1.72} \\
        \midrule
        Gibson~[\origref{73}]      &  43.79             & 31.66$\pm$1.05       & \textemdash     & \textemdash     & 46.67     & 33.74$\pm$1.21\\
        HM3D~[\origref{56}]        &  \textemdash       & \textemdash          & 43.79           & 31.66$\pm$1.05  & 26.33     & 21.70$\pm$1.03\\
        \bottomrule
        \end{tabular}
        }
    \label{tab:photorealism_comparison}
\end{table}

\subsection{Examples of Expert Demonstration}
\label{supp:examples_of_expert_demonstration}
Fig.~\ref{fig:supple_expert_demo1}-\ref{fig:supple_expert_demo7} illustrate the examples of expert demonstrations for 7 task types.
The agent has to solve the task in interactive environments by understanding the language instructions and planning the sequential and executable actions.

\subsection{Diverse Episodes with More Objects}
Each episode is generated based on the combination of task-relevant objects (\eg, \emph{put a `knife' on the `table.'}) and this indicates that more object classes can result in more object-diverse episodes (\eg, \emph{put a `potato' in a `fridge.'}). 
We observe that our \method provides more diverse episodes compared to the ALFRED benchmark~[\origref{61}] by enriching the number of object classes.
Here, we denote an episode whose combination of task-relevant objects does not overlap with the others by a unique episode.
We observe that our \method benchmark provides the $4,649$ unique episodes while ALFRED provides $2,522$ ones in the combined \textit{train} and \textit{valid} splits.
In addition, the ratio of the unique episodes among the total ones is 53.3\% in \method while it is 35.6\% in ALFRED.
This indicates that our \method benchmark provides not only a larger number of episodes but also more diverse combinations of episodes.

\subsection{Qualitative Comparison of Indoor Houses}
We provide a qualitative comparison of the indoor houses used in our \method and ALFRED~[\origref{61}] in the attached video files (\textit{video1.mp4}, \textit{video2.mp4}, and \textit{video3.mp4}).
We provide the agent's egocentric view on the left-hand side of a video and a top-down view with a red circle denoting the agent's corresponding current location.
While ALFRED's indoor house environments consist of single room types on a room scale, ours are provided on a house scale, featuring multiple rooms within a single house.
This implies that agents developed with our environments are enabled to perform instruction-following tasks that require navigating through multiple rooms.

\section{Details of State of the Art Models}
We provide details of state-of-the-art models in imitation learning and spatial map reconstruction, respectively.

\subsection{Imitation Learning}
The Seq2Seq~[\origref{61}] model encodes the visual input with the frozen backbone visual encoder.
The natural language goal and instructions are encoded with a bidirectional LSTM encoder to produce an embedding for each word.
Alongside the previous action, embeddings are passed as input to an LSTM cell to produce the current hidden state.
The action and corresponding mask are finally predicted using a hidden state.
MOCA~[\origref{62}] exploits separate branches for action prediction and object localization to better address different semantic understanding.
ABP~[\origref{33}] extends MOCA~[\origref{62}] by perceiving surrounding perception for a better understanding of environments with the enlarged field of view.

\subsection{Spatial Map Reconstruction}
HLSM~[\origref{5}] uses a hierarchical controller to bridge the gap between natural language instructions and agent executable actions.
The high-level controller predicts the next subgoal given the instruction and the map, and then the low-level controller outputs a sequence of actions to achieve the subgoal.
FILM~[\origref{46}] utilizes a pre-designed template as a high-level action sequence.
It uses two submodules of BERT classifiers to predict the type of instruction and the arguments to fill in the template.
Finally, it uses a deterministic algorithm~[\origref{60}] for obstacle-free path planning.
LLM-Planner~[\origref{63}] leverages large language model to generate subgoal sequence with a few examples.
To enhance LLMs planning accuracy, it updates plans that are physically grounded in the environment.
CAPEAM~[\origref{34}] uses context-aware planning to plan a subgoal sequence and conduct the respective subgoal with the corresponding detailed planners.
It also uses additional memory to prevent the interaction of inappropriate objects.

\section{Extended Quantitative Results}
\label{sec:supp_extended_qauntitative}
We present experiment results with path-length-weighted success rate and goal condition (\ie, PLWSR and PLWGC) over multiple runs in Table~\ref{tab:ext-baseline-result-1} and~\ref{tab:ext-baseline-result-2}.

\newcommand{\mcp}[1]{\multicolumn{#1}{c@{\hspace{30pt}}}}
\definecolor{Gray}{gray}{0.90}
\newcolumntype{a}{>{\columncolor{Gray}}r}
\newcolumntype{b}{>{\columncolor{Gray}}c}

\begin{table*}[t!]
    \caption{
        \textbf{Task and Goal-Condition Success Rate (valid split).}
        Path-length-weighted (PLW) metrics are given in parentheses for each value.
        We report mean and stndadrd deviation over multiple runs.
        $^\dagger$Authors' implementation as the code is not publicly available.
    }
    \vspace{-.75em}
    \centering
    \resizebox{1.00\textwidth}{!}{
        \begin{tabular}{@{}llaarr@{}}
            \toprule
            \multirow{3}{*}{Learning}& \multirow{3}{*}{Model}
                             & \mcp{4}{\textbf{Validation}} \\
                             \cmidrule{3-6}
                             & & \mcc{2}{\textit{Seen}} & \mcc{2}{\textit{Unseen}}\\
                             & & \multicolumn{1}{b}{Success Rate} & \multicolumn{1}{b}{Goal Condition} 
                             & \multicolumn{1}{c}{Success Rate} & \multicolumn{1}{c}{Goal Condition} \\
            
            \cmidrule{1-6}
            
            \multirow{3}{*}{\parbox[c]{2cm}{\centering\textbf{Imitation \\Learning}}} & {Seq2Seq~[\origref{61}]}     & $0.77\pm0.06$ ($0.47\pm0.06$)    & $6.93\pm0.06$ ($4.73\pm0.06$)    & $0.00\pm0.00$ ($0.00\pm0.00$)   & $4.03\pm0.06$ ($2.50\pm0.00$)  \\
            & {MOCA~[\origref{62}]}      & $12.64\pm0.12$ ($8.35\pm0.16$)     & $20.95\pm0.18$  ($13.43\pm0.16$)   & $1.44\pm0.05$ ($0.56\pm0.06$)     & $6.76\pm0.04$ ($3.64\pm0.06$)    \\
            & {ABP$^\dagger$~[\origref{33}]}               & $24.71\pm0.05$ ($15.49\pm0.34$)    & $33.80\pm0.14$  ($23.27\pm0.32$)   & $4.22\pm0.05$ ($1.70\pm0.08$)     & $11.71\pm0.27$ ($5.42\pm0.13$)   \\
            
            \cmidrule{1-6}
            
            \multirow{4}{*}{\parbox[c]{2cm}{\centering\textbf{Spatial\\Map\\Reconst.}}} & {HLSM~[\origref{5}]}      & $4.23\pm0.08$ ($0.72\pm0.08$)            & $9.14\pm0.09$  ($2.67\pm0.06$)           & $1.08\pm0.14$ ($0.19\pm0.03$)           & $6.12\pm0.23$ ($1.52\pm0.02$)         \\
            & {FILM~[\origref{46}]}                & $7.08\pm0.28$ ($1.87\pm0.11$)      & $11.93\pm0.23$  ($4.82\pm0.15$)     & $4.44\pm0.17$ ($1.25\pm0.10$)     & $9.26\pm0.13$ ($3.84\pm0.11$)   \\
            & {LLM-Planner$^\dagger$~[\origref{63}]}   & $5.80\pm0.19$ ($1.51\pm0.03$)      & $11.69\pm0.35$  ($4.76\pm0.07$)     & $3.33\pm0.22$ ($0.96\pm0.05$)     & $8.29\pm0.19$ ($3.49\pm0.09$)   \\
            & {CAPEAM$^\dagger$~[\origref{34}]}           & $13.45\pm0.05$ ($3.43\pm0.06$)      & $18.16\pm0.27$  ($4.50\pm0.05$)     & $4.92\pm0.22$ ($1.22\pm0.03$)     & $9.47\pm0.23$ ($1.79\pm0.04$)   \\

            \bottomrule
        \end{tabular}
    }
    
    \label{tab:ext-baseline-result-1}
\end{table*}

\begin{table*}[t!]
    \caption{
        \textbf{Task and Goal-Condition Success Rate (test split).}
        Path-length-weighted (PLW) metrics are given in parentheses for each value.
        We report mean and stndadrd deviation over multiple runs.
        $^\dagger$Authors' implementation as the code is not publicly available.
    }
    \vspace{-.75em}
    \centering
    \resizebox{1.00\textwidth}{!}{
        \begin{tabular}{@{}llaarr@{}}
            \toprule
            \multirow{3}{*}{Learning}& \multirow{3}{*}{Model}
                             & \mcp{4}{\textbf{Test}} \\
                             \cmidrule{3-6}
                             & & \mcc{2}{\textit{Seen}} & \mcc{2}{\textit{Unseen}}\\
                             & & \multicolumn{1}{b}{Success Rate} & \multicolumn{1}{b}{Goal Condition} 
                             & \multicolumn{1}{c}{Success Rate} & \multicolumn{1}{c}{Goal Condition} \\
            
            \cmidrule{1-6}
            
            \multirow{3}{*}{\parbox[c]{2cm}{\centering\textbf{Imitation \\Learning}}} & {Seq2Seq~[\origref{61}]}     & $1.10\pm0.00$ ($0.05\pm0.01$)   & $6.60\pm0.00$ ($5.00\pm0.00$)   & $0.00\pm0.00$ ($0.00\pm0.00$) & $3.50\pm0.00$ ($2.80\pm0.00$) \\
            & {MOCA~[\origref{62}]}      & $14.11\pm0.03$ ($9.20\pm0.05$)    & $22.84\pm0.04$ ($16.42\pm0.04$)  & $0.62\pm0.08$ ($0.35\pm0.05$)   & $5.14\pm0.08$ ($3.39\pm0.06$) \\
            & {ABP$^\dagger$~[\origref{33}]}               &  $27.44\pm0.40$ ($16.96\pm0.16$)   & $35.81\pm0.23$ ($24.57\pm0.19$)   & $3.54\pm0.23$ ($1.51\pm0.08$) & $10.57\pm0.22$ ($5.59\pm0.10$) \\
            \cmidrule{1-6}
            
            \multirow{4}{*}{\parbox[c]{2cm}{\centering\textbf{Spatial\\Map\\Reconst.}}} & {HLSM~[\origref{5}]}      & $6.27\pm0.04$ ($0.88\pm0.10$)             & $10.44\pm0.13$ ($2.78\pm0.10$)          & $0.49\pm0.16$ ($0.08\pm0.03$)            & $4.28\pm0.13$ ($1.37\pm0.16$) \\
            & {FILM~[\origref{46}]}                & $8.79\pm0.07$ ($2.36\pm0.01$)    & $13.03\pm0.08$ ($5.58\pm0.08$)      & $2.15\pm0.18$ ($0.56\pm0.04$)      & $6.56\pm0.15$ ($3.16\pm0.05$) \\
            & {LLM-Planner$^\dagger$~[\origref{63}]}   & $8.16\pm0.20$ ($2.20\pm0.06$)    & $13.20\pm0.13$ ($5.72\pm0.06$)      & $1.90\pm0.13$ ($0.57\pm0.04$)      & $6.33\pm0.02$ ($3.09\pm0.04$) \\
            & {CAPEAM$^\dagger$~[\origref{34}]}           & $15.61\pm0.15$ ($3.68\pm0.09$)    & $20.22\pm0.11$ ($5.39\pm0.09$)      & $2.87\pm0.13$ ($0.84\pm0.02$)      & $7.36\pm0.07$ ($2.01\pm0.03$) \\
            
            \bottomrule
        \end{tabular}
    }
    
    \label{tab:ext-baseline-result-2}
\end{table*}

\section{Map Reconstruction Strategy}
\label{sec:supp_map_recon}
We provide a more detailed analysis of the challenges in recognizing narrow passages (\eg, doors, aisles, \etc).
Our observation is as follows: failure to recognize narrow navigable pathways leads an agent to \textit{stuck} within the initial room.

\begin{figure}[t!]
    \centering
    \includegraphics[width=0.75\linewidth]{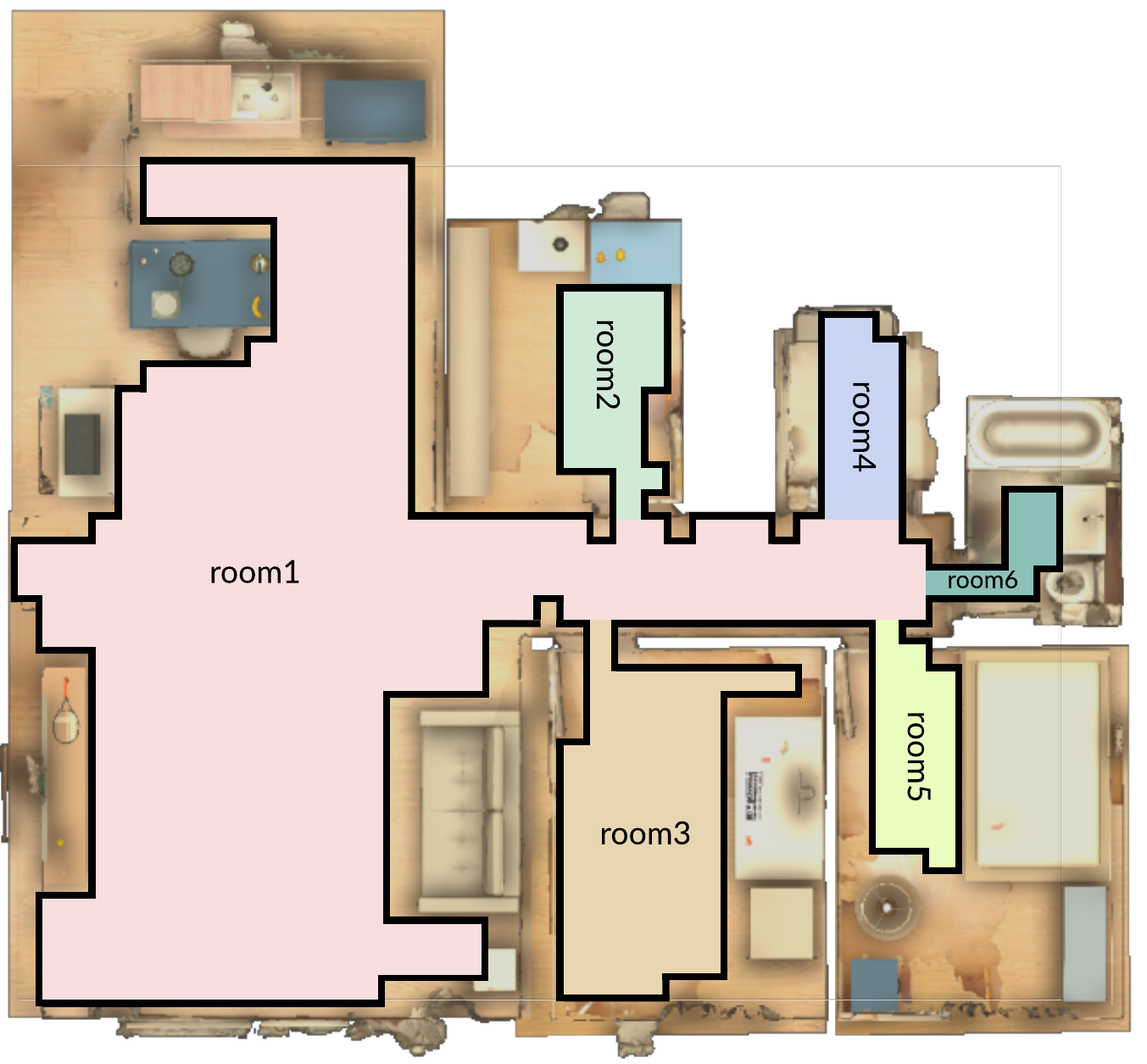}
    \vspace{-1em}
    \caption{
        \textbf{Top-down view of a scene with labeling.}
        We select one of the largest scene in \textit{valid unseen} fold and annotate the space from room 1 to room 6 based on the door gap for further analysis.
        The area inside the outer black line is a navigable area.
    }
    \label{fig:supple_scene_labeled}
\end{figure}

To quantify this challenge, we establish a criterion for \textit{leaving} a room by taking $1$ step (\ie $0.25$ meter) further from the entrance of the room where the agent was initiated.
Fig.~\ref{fig:supple_scene_labeled} represents the top-down view of one of \textit{valid unseen} scenes labeled for each space from room $1$ to room $6$.
We observe that only $5.3\%$ of the agents, from the total episodes conducted on the scene, left one room to another for further exploration.
We also noticed that above mentioned \textit{leaving} occurred within a short span, not exceeding $77$ steps.
This implies that if an agent initially overlooks narrow spaces, they will be mistaken for walls when viewed at an oblique angle.

\begin{figure}[t!]
\centering
    \begin{subfigure}{0.45\linewidth}
        \centering
        \includegraphics[width=0.9\linewidth]{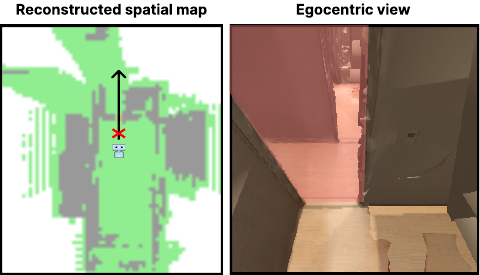}
    \end{subfigure}
    \begin{subfigure}{0.45\linewidth}
        \centering
        \includegraphics[width=0.9\linewidth]{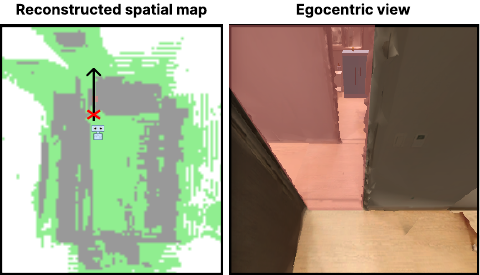}
    \end{subfigure}
    \caption{
        \textbf{Reconstructed spatial map and egocentric view.}
        A reconstructed spatial map is presented on the left-hand side and an egocentric view of the agent is presented on the right-hand side on each figures.
        The green area denotes a predicted navigable area and the dark gray area denotes a predicted obstacle.
        Agent fails to recognize narrow navigable space (highlighted in red on the right-hand side of the each figure).
    }    
    \label{fig:supp_spatialmap}
    \vspace{-.5em}
\end{figure}

We provide qualitative examples in Fig.~\ref{fig:supp_spatialmap}, illustrating failure cases in reconstructing narrow passages.
We observe that an agent that adopted a spatial map fails to reconstruct its surroundings, fails to recognize narrow doorways, gets \textit{stuck} in a single room, and eventually, fails to complete a task.

We further explore the likelihood of an agent, under a random navigation policy, to be positioned where it can directly observe narrow spaces or door gaps to construct a semantic map accurately.
We assess the likelihood within $100$ steps (\ie a slight buffer extended to $77$ steps).
In detail, we consider the agent's position, viewing direction, and the horizontal angle of the head to measure the likelihood.
As a result, the agent has a $6.9\%$ likelihood of aligning with and facing the exit passage to recognize it, similar to the empirical result of the agent leaving the room where it was initiated (\ie, $5.3\%$).
This may imply that not being positioned correctly to see doors shortly after its deployment increases the chance of getting stuck in a room.
This consideration may not have been necessary in the ALFRED benchmark~[\origref{61}], which consists of single rooms.
However, since \method is composed of multiple rooms, `Spatial Map Reconst.' baselines may achieve much lower performance compared to [\origref{61}].

\section{Qualitative examples for domain adaptation}

\begin{figure}[t!]
    \centering
    \includegraphics[width=0.75\linewidth]{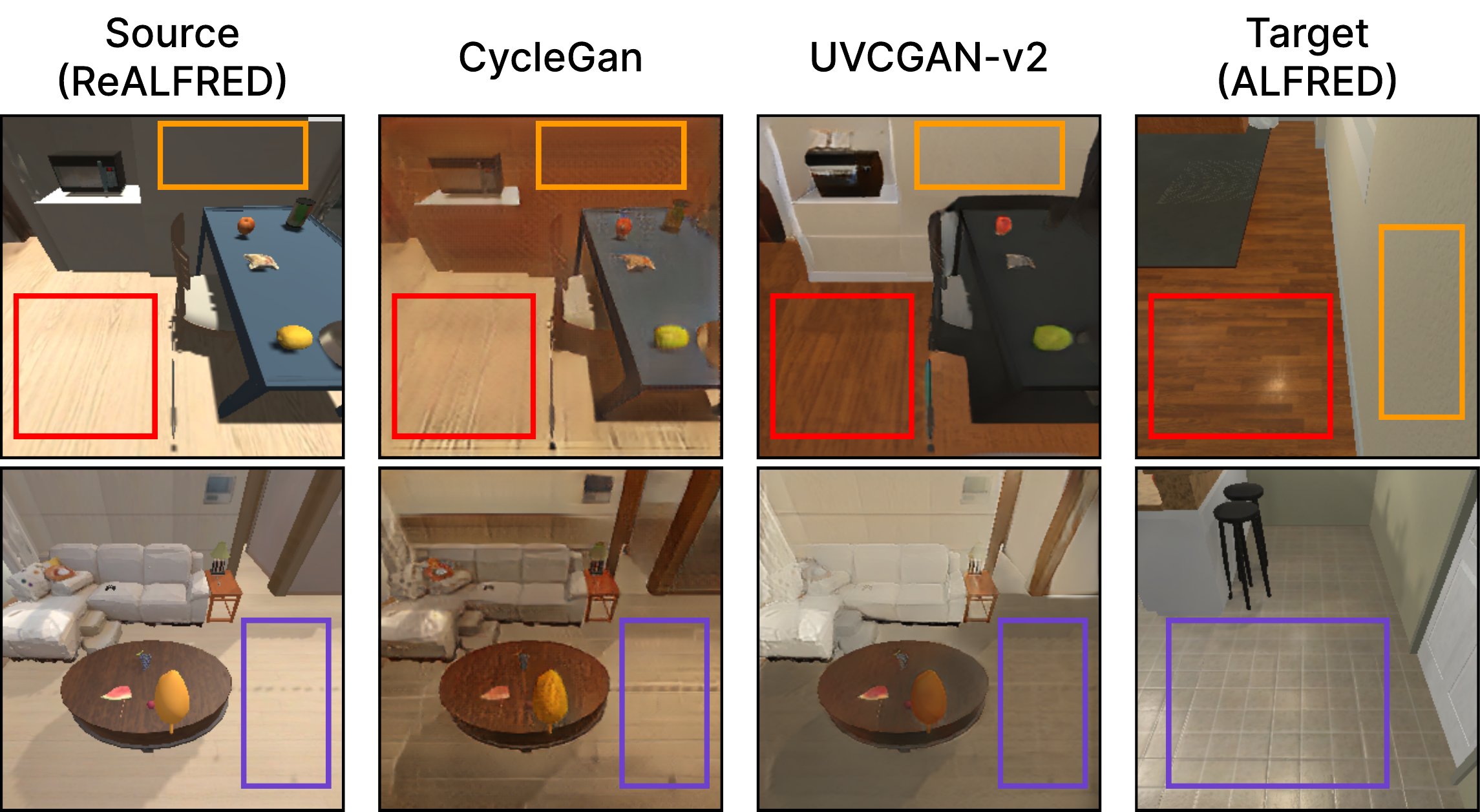}
    \caption{
        \textbf{Qualitative examples of the real-to-sim domain adaptation.}
        In each column, we provide source image, domain adapted image for the source image, and an image from target domain.
    }
    \label{fig:supp_gan}
    \vspace{-.5em}
\end{figure}

We provide qualitative examples of real-to-sim domain adaptation in Fig.~\ref{fig:supp_gan}.
`Source' column denotes a visual frame from the \method benchmark, `CycleGan' column denotes an adpated visual frame with CycleGan~[\origref{79}], `UVCGAN-v2' column denotes an adapted visual frame with UVCGAN-v2~[\origref{70}], and `Target' column denotes an image from ALFRED target domain image.
The advantage of real-to-sim domain adaptation is to make an agent feel at home, reducing the visual domain gap.
We expect a domain adapted image to resemble some characteristics that are well represented in the target domain, where sim2real agent is trained.

We begin our examination with an example in the first row.
We observe that the flooring, dominating the image frame, is adapted to resemble the flooring that frequently appears in the target domain (highlighted with \redbox).
We also notice that the image generated with CycleGan adapts the color of the wall to brown, while the image generated with UVCGAN-v2 adapts the wall to white tone (highlighted with \orangebox).

We now examine an example in the second row.
We observe that the flooring, dominating the image frame as in the first example, is generated to resemble checkerboard tiles which are represented in target domain (highlighted with \violetbox).

\begin{figure}[t]
    \centering
    \includegraphics[width=0.98\linewidth]{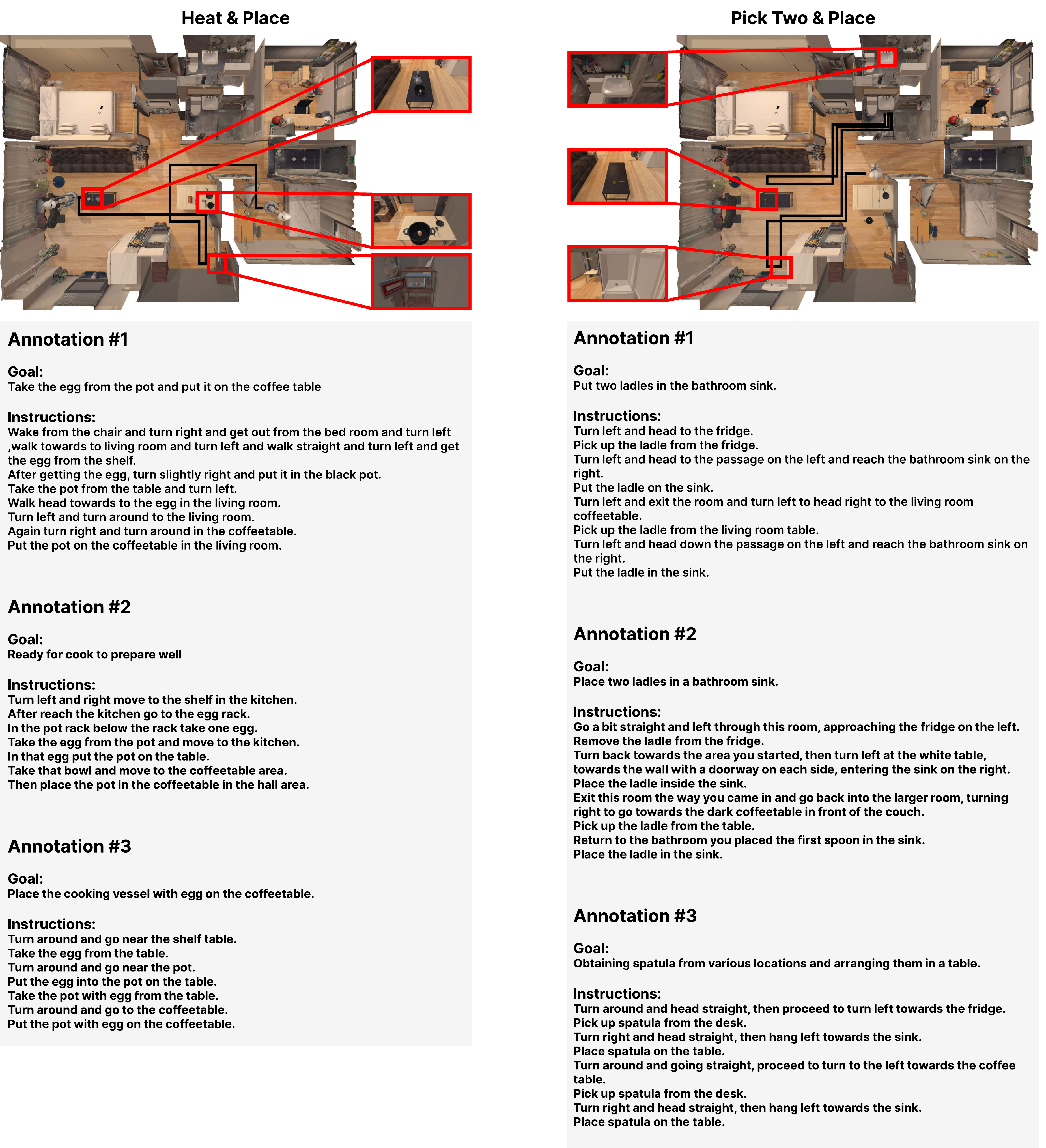}
    \vspace{-.5em}
    \caption{
        \textbf{Examples of expert demonstration and human annotation (`Heat \& Place' on the left and `Pick Two \& Place' on the right).}
        We provide examples of expert demonstration for tasks `Heat \& Place' and `Pick Two \& Place.'
        The black lines denote the expert's trajectories, and several egocentric views are presented alongside a top-down view of the scene.
    }
    \label{fig:supple_expert_demo1}
\end{figure}

\begin{figure}[t!]
    \centering
    \includegraphics[width=0.98\linewidth]{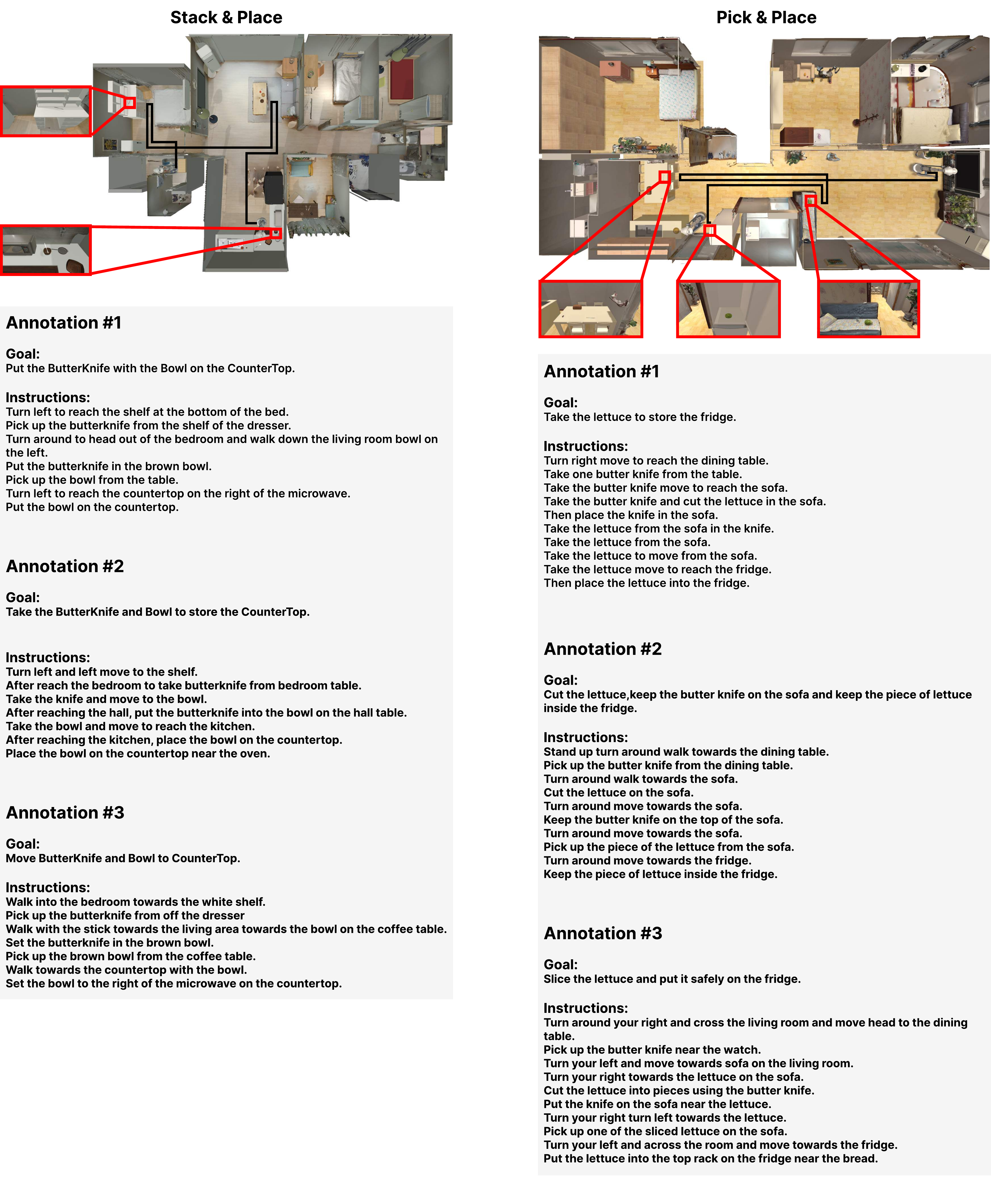}
    \vspace{-1em}
    \caption{
        \textbf{Examples of expert demonstration and human annotation (`Stack \& Place' on the left and `Pick \& Place' on the right).}
        We provide examples of expert demonstration for tasks `Stack \& Place' and `Pick \& Place.'
        The black lines denote the expert's trajectories, and several egocentric views are presented alongside a top-down view of the scene.
    }
    \label{fig:supple_expert_demo2}
\end{figure}

\begin{figure}[t!]
    \centering
    \includegraphics[width=0.98\linewidth]{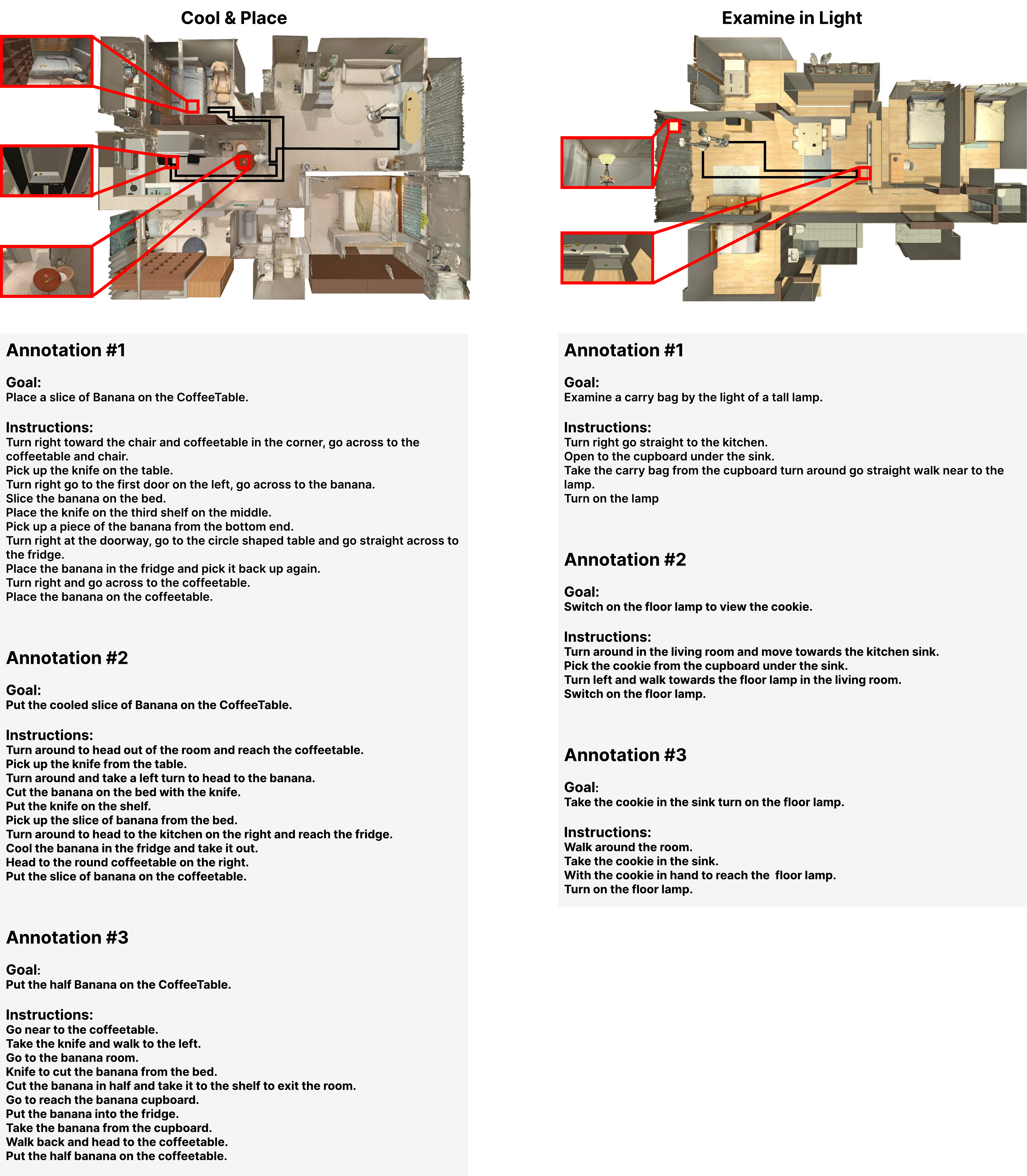}
    \vspace{-1em}
    \caption{
        \textbf{Examples of expert demonstration and human annotation (`Cool \& Place' on the left and `Examine in Light' on the right).}
        We provide examples of expert demonstration for tasks `Cool \& Place' and `Examine in Light.'
        The black lines denote the expert's trajectories, and several egocentric views are presented alongside a top-down view of the scene.
    }
    \label{fig:supple_expert_demo3}
\end{figure}

\begin{figure}[t!]
    \centering
    \includegraphics[width=0.55\linewidth]{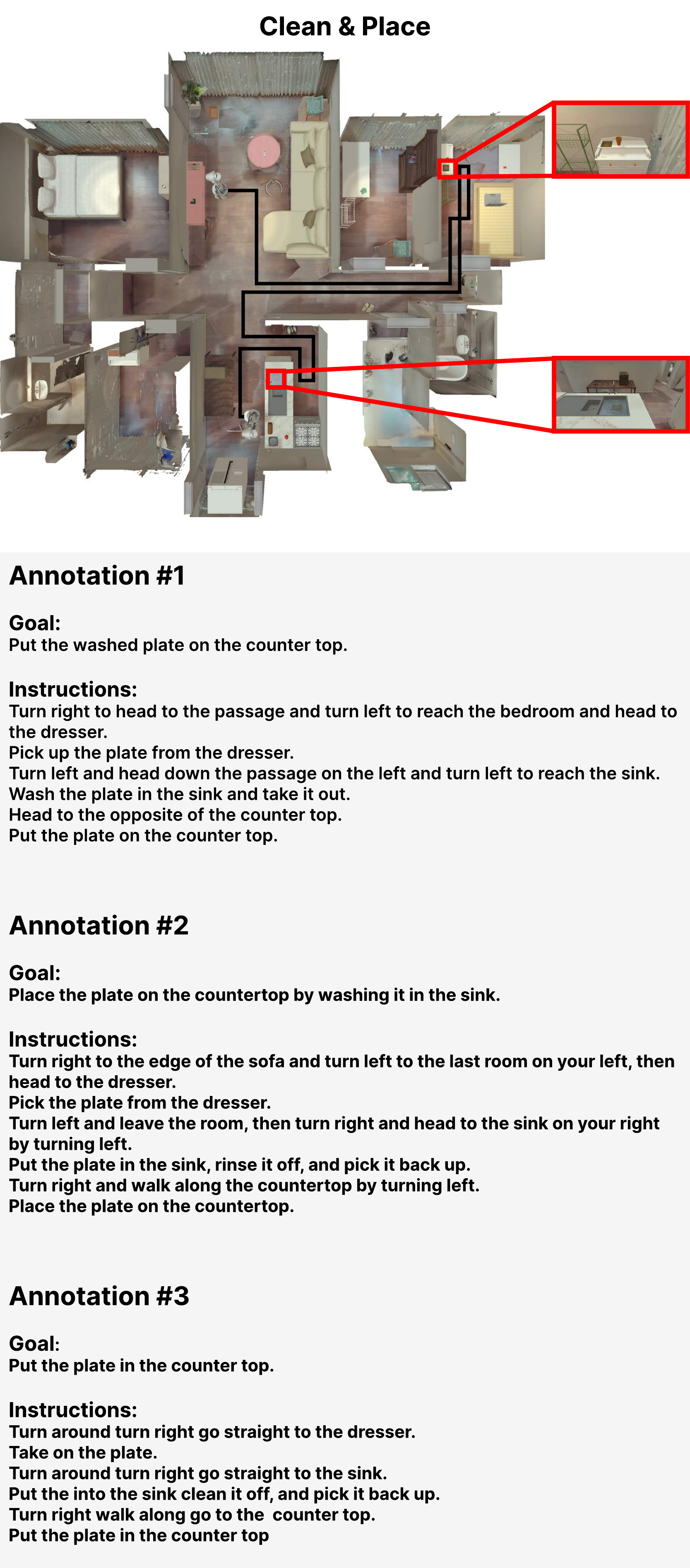}
    \vspace{-.25em}
    \caption{
        \textbf{Example of expert demonstration and human annotation (`Clean \& Place').}
        We provide an example of an expert demonstration for task `Clean \& Place.'
        The black line denotes the expert's trajectory, and several egocentric views are presented alongside a top-down view of the scene.
    }
    \label{fig:supple_expert_demo7}
\end{figure}

\end{document}